\documentclass{article}

\usepackage{microtype}
\usepackage{graphicx}
\usepackage{subcaption}
\usepackage{booktabs} 
\usepackage{lineno}
\usepackage{enumitem}

\usepackage{times}
\usepackage{latexsym}
\usepackage{amsmath}
\usepackage{amssymb}
\usepackage{booktabs}
\usepackage{multirow}
\usepackage{makecell}
\usepackage[table]{xcolor}
\usepackage[T1]{fontenc}
\usepackage{pifont}
\usepackage{array}
\usepackage[utf8]{inputenc}
\usepackage{multirow}
\usepackage{microtype}
\usepackage{placeins}

\usepackage{hyperref}

\usepackage[preprint]{icml2026}

\usepackage{amsmath}
\usepackage{amssymb}
\usepackage{mathtools}
\usepackage{amsthm} 

\usepackage[capitalize,noabbrev]{cleveref}

\theoremstyle{plain}

\theoremstyle{definition}

\theoremstyle{remark}

\usepackage[textsize=tiny]{todonotes}

\icmltitlerunning{Rethinking Selective Knowledge Distillation}

\newcommand\footnoteONLYtext[1]{
    \let\svthefootnote\thefootnote
    \let\thefootnote\relax\footnotetext{\hspace{-1ex}#1}
    \let\thefootnote\svthefootnote
}

\newcommand{\selector}{\textsc{SE-KD}}
\newcommand{\triaxselector}{\textsc{SE-KD}$_{\text{3X}}$}

\newcommand{\cmark}{\ding{51}} 
\newcommand{\xmark}{\ding{55}} 

\begin{document}

\twocolumn[
  \icmltitle{Rethinking Selective Knowledge Distillation}



  \icmlsetsymbol{equal}{*}

  \begin{icmlauthorlist}
    \icmlauthor{Almog Tavor}{yyy}
    \icmlauthor{Itay Ebenspanger}{yyy,equal}
    \icmlauthor{Neil Cnaan}{yyy,equal}
    \icmlauthor{Mor Geva}{yyy}
  \end{icmlauthorlist}

  \icmlaffiliation{yyy}{Blavatnik School of Computer Science and AI, Tel Aviv University}

  \icmlcorrespondingauthor{Almog Tavor}{almogt@mail.tau.ac.il}

  \icmlkeywords{Machine Learning, ICML}

  \vskip 0.3in
]



\printAffiliationsAndNotice{\icmlEqualContribution}

\begin{abstract}
Growing efforts to improve knowledge distillation (KD) in large language models (LLMs) replace dense teacher supervision with selective distillation, which uses a subset of token positions, vocabulary classes, or training samples for supervision.
However, it remains unclear which importance signals, selection policies, and their interplay are most effective. In this work, we revisit \textit{where} and \textit{how} to distill in autoregressive LLMs. 
We disentangle selective KD along the position, class, and sample axes and systematically compare importance signals and selection policies. Then, guided by this analysis, we identify underexplored opportunities and introduce student-entropy-guided position selection (\selector{}). 
Across a suite of benchmarks, \selector{} often improves accuracy, downstream task adherence, and memory efficiency over dense distillation.
Extending this approach across the class and sample axes (\triaxselector{}) yields complementary efficiency gains that make offline teacher caching feasible. In practice, this reduces wall time by 70\% and peak memory by 18\%, while cutting storage usage by 80\% over prior methods without sacrificing performance.
\end{abstract}

\section{Introduction}
Large language models (LLMs) achieve state-of-the-art results across diverse tasks, but their size makes them expensive to serve and difficult to adapt.
Knowledge distillation \citep[KD;][]{hinton2015distillation} addresses this by training a smaller student model to imitate a larger teacher.
For autoregressive LLMs, this is typically done by matching the teacher's next-token distribution at every position of the training sequence.

However, applying knowledge distillation at every token position is often suboptimal due to the uniform supervision across all positions.
Recent studies demonstrate that performance can be improved by selecting or reweighting positions for KD based on signals such as student cross-entropy \citep{wang2021selectivekd}, teacher uncertainty \citep{zhong2024atkd,huang2025selectkd}, and teacher-student discrepancy \citep{xie2025adakd}.
Yet, it remains unclear which token-importance signals most reliably identify positions that benefit from logit-based distillation in LLMs, and how different position-selection policies interact with these signals to shape an effective distillation curriculum.

\begin{figure*}[t]
\setlength{\belowcaptionskip}{-8pt}
    \centering
    \includegraphics[width=\linewidth]{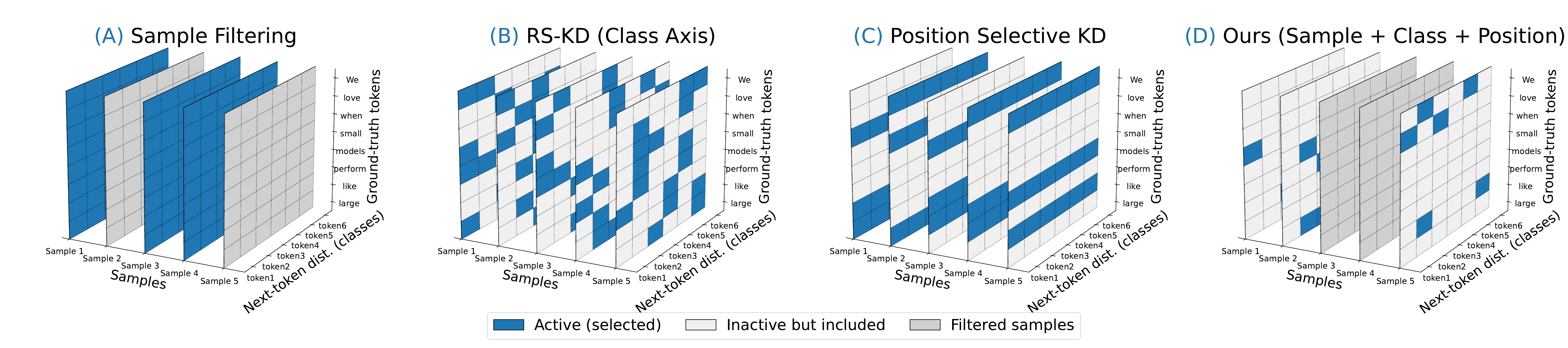}
    \caption{Illustration of  three key selection axes for knowledge distillation:
    \textbf{(A)} sample selection,
    \textbf{(B)} class sampling (RS-KD),
    \textbf{(C)} position-selective KD,
    and \textbf{(D)} our combined approach, \triaxselector{}, which integrates sample, class, and position selection.
    Blue cells denote active (selected) supervision, light gray indicates inactive but included elements, and dark gray denotes filtered samples.
    }
    \label{fig:3-axis-se-kd}
\end{figure*}

In this work, we revisit \textit{where} and \textit{how} to apply teacher supervision in knowledge distillation for autoregressive LLMs.
We first disentangle selective KD into five design axes: the alignment criterion, positions, classes, samples, and features.
Within this framework, we focus on 3 key selection axes (Fig.~\ref{fig:3-axis-se-kd})---positions, classes, and samples---and systematically analyze:
(i) the choice of position-importance signal, comparing uncertainty and discrepancy-based measures such as entropy and teacher--student KL;
(ii) the policy used to convert these signals into selective supervision, e.g., top-$k$ selection, curriculum learning, and stochastic allocation under fixed budgets; and
(iii) how position selection interacts with sparsification along the class and sample axes.

Motivated by gaps revealed by this analysis, we identify two underexplored opportunities:
(1) the use of \emph{student entropy} as a position-importance signal, and
(2) joint selection across multiple axes.
We address these gaps by introducing a student-entropy-guided position-selective KD method, called \selector{}, and its 3-axis variant \triaxselector{}, which applies selection over samples, positions, and classes (Fig.~\ref{fig:3-axis-se-kd}D).

Through experiments across a broad suite of benchmarks, covering 9 importance signals, 5 selection policies, and 6 KD baselines, we find that student-uncertainty-based position selection reliably identifies high-value tokens for distillation.
Selecting the top-20\% positions based on student-entropy yields a consistent improvement in average evaluation accuracy (64.8 vs. 64.4 for Full KD) and perplexity (6.9 vs. 7.3), while requiring supervision on only a fraction of token positions.
These gains come with some reduction in calibration (0.273 $\rightarrow$ 0.276), yet substantially reduce computational and memory overhead, as fewer teacher and student logits need to be computed.

Next, we evaluate \selector{} and \triaxselector{} on two additional settings of on-policy distillation \citep{agarwal2024onpolicy}, where supervision is applied to student-generated trajectories, and task-specific distillation, focusing on math reasoning.
We find that our approach remains competitive in both settings, suggesting that our method generalizes beyond a single distillation regime.

Finally, we analyze the efficiency gains of position-selective KD when combined with sample selection and class sampling. On 80M token distillation, student-entropy-based sample selection reduces total wall time by 70\%,
and class-sampled offline caching becomes feasible in practice, cutting storage by 99.96\%, while maintaining performance.

In conclusion, our work makes the following contributions:
\begin{itemize}
[leftmargin=*,topsep=0pt,itemsep=3pt,parsep=0pt]
    \item We propose a general, theoretical framework for selective KD that organizes prior methods and highlights unexplored variants.
    
    \item We introduce new selective KD variants, \selector{} and \triaxselector{}, guided by \emph{student entropy}. We show that these variants provide an often best-performing signal for KD, outperforming prior position-importance metrics across position-selective, and yielding the strongest gains in accuracy, while preserving downstream task adherence.
    
    \item We show that \selector{}, combined with class and sample selection, substantially improves distillation efficiency via offline teacher cache and selection-aware implementations (selective LM head and chunked entropy computation), reducing runtime, peak memory, and storage.
\end{itemize}
We release our code at: \url{https://github.com/almogtavor/SE-KD3x}.

\section{Related Work}

\paragraph{KD with Position Selection}
A prominent line of work has explored ways to improve KD by selectively apply supervision at only part of the sequence positions.
\citet{wang2021selectivekd} selected the top \(k\%\) positions with the highest student cross-entropy using both batch-local selection and global-level selection (GLS).
More recently, \citet{huang2025selectkd} proposed down-weighting positions whose student proposals are not supported by the teacher.
In parallel, token-adaptive frameworks dynamically adjust token-level supervision based on teacher-student distribution discrepancy \citep{xie2025adakd}.
These approaches focus on a single selection heuristic or setting, broadly following an ``80/20'' intuition: a small fraction of high-entropy ``fork'' positions may carry much of the distillation signal \citep{wang2025highentropy}.
Our work extends these efforts by providing a unified comparison of position-selection strategies and metrics under a common distillation setup, isolating which signals reliably identify informative positions across tasks.

\paragraph{Curriculum Learning} For chain-of-thought distillation, \citet{feng2024kpod} learned position-importance weights and used a curriculum that expands supervision from easier to harder reasoning steps under a given budget. Inspired by this work, we incorporate curriculum in two ways: (1) our student-entropy selection induces an implicit curriculum as supervised positions adapt during training; and (2) we evaluate an explicit curriculum-style position-selection method.

\paragraph{Uncertainty-Guided Position-Weighting KD}
Previous work showed that uncertainty-weighted distillation can improve reliability and calibration \citep{guo2024entropykd}.
Recently, Adaptive-Teaching KD \citep[AT-KD;][]{zhong2024atkd}
built on Decoupled KD \citep{zhao2022decoupled} and routes token-level supervision using the teacher's gold-label probability,
\(1 - p_t(y_t)\), where \(p_t(y_t)\) is the teacher probability assigned to the ground-truth next token.
Per batch, AT-KD ranks positions by this uncertainty score and splits them into easy and hard tokens, skipping the target-class KL term on easy tokens while emphasizing diversity on hard tokens.
Unlike prior approaches that incorporate uncertainty through position-wise loss reweighting, our method uses uncertainty solely as a ranking signal for explicit selection.

\paragraph{KD with Class Sampling}
A complementary line of work has focused on reducing  distillation cost by sparsifying the teacher's output distribution.
Deterministic top-$k$ or percentile truncation of teacher logits \citep{raman2023slim, shum2024first} reduces compute and storage costs but discards tail mass, inducing biased gradient estimates and miscalibrated students.
Random-Sampling KD \citep[RS-KD; ][]{anshumann2025sparse} replaced truncation with importance sampling to provide unbiased gradient estimates and improved calibration.
These works focus on class sampling, which is one of the selection axes that we study.

\paragraph{KD with Sample Selection \& Weighting}
Distillation efficiency can also be improved by reducing the number of teacher queries. 
For example, UNIX \citep{xu2020unix} uses uncertainty-aware sampling to focus distillation on informative samples.
Other work focused on the sample selection to improve accuracy. Entropy-based adaptive KD reweighs the KD loss by prioritizing \textit{samples} according to  
the entropy of the teacher and student \citep{su2023eakd}.
More recently, Difficulty-Aware Knowledge Distillation (DA-KD) \citep{he2025dakd} explicitly measures sample difficulty via the discrepancy between teacher and student cross-entropy losses, defined as the CE ratio, $\mathcal{L}^{\mathrm{CE}}_{\text{student}}(x)/\mathcal{L}^{\mathrm{CE}}_{\text{teacher}}(x)$, 
and utilizes this score for difficulty-aware stratified sampling, so that distillation focuses on hard but informative examples while maintaining data diversity.
Our work considers both teacher–student and student-only based sample selection.

\begin{table*}[th!]
\centering
\caption{\textbf{Overview of selective KD methods with selection-axis membership.}
The columns \textbf{Pos}, \textbf{Cls}, and \textbf{Smp} indicate whether a method applies selection/sparsification along the position, class, or sample axes, respectively.
{\textcolor[HTML]{34A853}{\cmark}} denotes that the method explicitly acts on that axis, while {\textcolor[HTML]{ff0000}{\xmark}} indicates it does not.
We highlight our proposed student-entropy variants in green.}
\footnotesize
\setlength{\tabcolsep}{4pt}
\renewcommand{\arraystretch}{1.1}
\begin{tabular}{
p{1.7cm} p{4.6cm} p{7.4cm}
>{\centering\arraybackslash}p{0.55cm}
>{\centering\arraybackslash}p{0.55cm}
>{\centering\arraybackslash}p{0.55cm}
}
\toprule
& \textbf{Method} & \textbf{Description} &
\textbf{Pos} & \textbf{Cls} & \textbf{Smp} \\
\midrule

\multirow{5}{=}{\textbf{Alignment criterion}} 
& Full KD \citep{hinton2015distillation} 
& KL/CE on all positions (Eq.~\ref{eq:full-kd})
& \textcolor[HTML]{ff0000}{\xmark} & \textcolor[HTML]{ff0000}{\xmark} & \textcolor[HTML]{ff0000}{\xmark} \\
& Decoupled KD~\citep{zhao2022decoupled} 
& Reweighs target vs. non-target terms in the KL loss
& \textcolor[HTML]{ff0000}{\xmark} & \textcolor[HTML]{ff0000}{\xmark} & \textcolor[HTML]{ff0000}{\xmark} \\
& AT-KD~\citep{zhong2024atkd} 
& Routes positions into easy/hard buckets with separate KL terms using teacher's gold-label (\(y_t\)) probability \(1 - p_t(y_t)\)
& \textcolor[HTML]{34A853}{\cmark} & \textcolor[HTML]{ff0000}{\xmark} & \textcolor[HTML]{ff0000}{\xmark} \\
& Weighted KD \citep{guo2024entropykd} 
& Reweighs per-position KLD in the loss using \(w_t \propto u(t)\) 
& \textcolor[HTML]{34A853}{\cmark} & \textcolor[HTML]{ff0000}{\xmark} & \textcolor[HTML]{ff0000}{\xmark} \\

\midrule

\multirow{9}{=}{\textbf{Position-importance metric}} 
& Student CE~\citep{wang2021selectivekd} 
& Student cross-entropy \(\mathrm{CE}(y_t,q_t)\)
& \textcolor[HTML]{34A853}{\cmark} & \textcolor[HTML]{ff0000}{\xmark} & \textcolor[HTML]{ff0000}{\xmark} \\
& Teacher CE~\citep{zhong2024atkd} 
& Teacher cross-entropy \(\mathrm{CE}(y_t,p_t)\)
& \textcolor[HTML]{34A853}{\cmark} & \textcolor[HTML]{ff0000}{\xmark} & \textcolor[HTML]{ff0000}{\xmark} \\
& \cellcolor[HTML]{DFF6CC} Student entropy
& Student entropy \(H(q_t)\) 
& \textcolor[HTML]{34A853}{\cmark} & \textcolor[HTML]{ff0000}{\xmark} & \textcolor[HTML]{ff0000}{\xmark} \\
& Teacher entropy 
& Teacher entropy \(H(p_t)\)
& \textcolor[HTML]{34A853}{\cmark} & \textcolor[HTML]{ff0000}{\xmark} & \textcolor[HTML]{ff0000}{\xmark} \\
& KL / reverse-KL 
& Teacher--student discrepancy \(\mathrm{KL}(p_t\|q_t)\) / \(\mathrm{KL}(q_t\|p_t)\)
& \textcolor[HTML]{34A853}{\cmark} & \textcolor[HTML]{ff0000}{\xmark} & \textcolor[HTML]{ff0000}{\xmark} \\
& \multirow{2}{*}{\cellcolor[HTML]{DFF6CC} KL + student entropy} 
& Combined discrepancy and uncertainty ranking \(\,\mathrm{KL}(p_t\|q_t)+\,H(q_t)\)
& \textcolor[HTML]{34A853}{\cmark} & \textcolor[HTML]{ff0000}{\xmark} & \textcolor[HTML]{ff0000}{\xmark} \\
& CE ratio~\citep{he2025dakd}
& Teacher--student CE ratio \(r(t)=\mathrm{CE}(y_t,q_t)/\mathrm{CE}(y_t,p_t)\)
& \textcolor[HTML]{34A853}{\cmark} & \textcolor[HTML]{ff0000}{\xmark} & \textcolor[HTML]{ff0000}{\xmark} \\

& \cellcolor[HTML]{DFF6CC} CE ratio + student entropy
& Combined difficulty and uncertainty \(r(t)+ H(q_t)\)
& \textcolor[HTML]{34A853}{\cmark} & \textcolor[HTML]{ff0000}{\xmark} & \textcolor[HTML]{ff0000}{\xmark} \\

\midrule
\multirow{5}{=}{\textbf{Position-selection policy}} 
& Top-$k$\%
& Chooses the top-$k$\% positions according to $u(t)$ 
& \textcolor[HTML]{34A853}{\cmark} & \textcolor[HTML]{ff0000}{\xmark} & \textcolor[HTML]{ff0000}{\xmark} \\
& GLS \citep{wang2021selectivekd}  
& Top-$k$\% normalized across batches with global thresholds \(\tau\)
& \textcolor[HTML]{34A853}{\cmark} & \textcolor[HTML]{ff0000}{\xmark} & \textcolor[HTML]{ff0000}{\xmark} \\
& Curriculum learning \citep{feng2024kpod} 
& Shifts supervision from easy to hard positions with scheduled window
& \textcolor[HTML]{34A853}{\cmark} & \textcolor[HTML]{ff0000}{\xmark} & \textcolor[HTML]{ff0000}{\xmark} \\
& \cellcolor[HTML]{DFF6CC} Pos RS-KD / Pos RS-KD$^*$
& Stochastic estimator \(q(t)\propto w_t\) of Weighted/Full KD
& \textcolor[HTML]{34A853}{\cmark} & \textcolor[HTML]{ff0000}{\xmark} & \textcolor[HTML]{ff0000}{\xmark} \\

\midrule
\multirow{4}{*}{\makecell[l]{\textbf{Class} \\ \textbf{sampling}}}
& \multirow{4}{*}{RS-KD~\citep{anshumann2025sparse}}
& At position \(t\), sample with repetition \(U\) indices \(v_k \propto p_t(v)\). Let \(\mathcal{C}_t=\{v_k\}_{k=1}^U\) be the unique sampled indices.
Build a sparse teacher target \(\tilde p_t\) on \(\mathcal{C}_t\) (from sampled counts) and minimize
\(\sum_{v\in\mathcal{C}_t}\tilde p_t(v)\log(\tilde p_t(v) / q_t(v))\).
& \textcolor[HTML]{ff0000}{\xmark} & \textcolor[HTML]{34A853}{\cmark} & \textcolor[HTML]{ff0000}{\xmark} \\

\midrule
\textbf{Sample selection} 
& Top-\(\ell\%\) avg.\ student entropy~\citep{xu2020unix}  
& Selects samples using student entropy \(U_i=\frac{1}{L_i-1}\sum_t H(q_t)\)
& \textcolor[HTML]{ff0000}{\xmark} & \textcolor[HTML]{ff0000}{\xmark} & \textcolor[HTML]{34A853}{\cmark} \\

\bottomrule
\end{tabular}
\label{tab:method-overview}
\end{table*}

\section{A Framework for Selective Knowledge Distillation}
\label{sec:framework}

We propose a general framework for selective KD, which encapsulates existing approaches and highlights opportunities for extending them. We then outline key design choices involved in the implementation of our framework.

\paragraph{Problem Setup}
In knowledge distillation, a student model is trained to imitate a teacher model by minimizing the divergence between their next-token distributions over a set of inputs.
Let \(x=(x_1,\ldots,x_L)\) be an input sample of $L$ tokens and $\mathcal{V}$ the shared vocabulary of the teacher and student.
At each position \(t\in\{1,\ldots,L-1\}\), the teacher and student define next-token distributions
\(p_t(\cdot)=p(\cdot\mid x_{\le t})\) and \(q_t(\cdot)=q(\cdot\mid x_{\le t})\) over $\mathcal{V}$, respectively.

The standard non-selective form, dubbed Full KD, optimizes a mixture of the teacher--student KL divergence and the ground-truth cross-entropy (CE), averaged over token positions and training samples.
For a given sample \(i\), the distillation loss \(\ell_{\mathrm{KD}}^{(i)}(t)\) at position $t$ is defined as
\begin{equation}
\label{eq:full-kd}
\ell_{\mathrm{KD}}(t)
=\lambda\,\mathrm{KL}\!\left(p_t \,\|\, q_t\right)
+ (1-\lambda)\,\mathrm{CE}(y_t, q_t),
\end{equation}
and for a training set \(\mathcal{D}\) the overall objective is
\begin{equation}
\mathcal{L}_{\mathrm{KD}}
=\frac{1}{|\mathcal{D}|}\sum_{i=1}^{|\mathcal{D}|}
\frac{1}{L_i-1}\sum_{t=1}^{L_i-1}\ell_{\mathrm{KD}}^{(i)}(t),
\end{equation}
where $\ell_{\mathrm{KD}}^{(i)}(t)$ is the loss at position $t$ of sample $i$, and $L_i$ is the length of sample $i$.

Selection therefore can be applied over three different axes: \textit{classes} at a specific position, \textit{positions} of a given sample, and \textit{samples} in the training set.
Let \(\mathrm{KL}_{\mathcal{C}_t^{(i)}}\) denote the KL divergence computed over a subset of classes
\(\mathcal{C}_t^{(i)} \subseteq \mathcal{V}\)
(where \(\mathcal{C}_t^{(i)}=\mathcal{V}\) for Full KD).
Moreover, let \(m_t^{(i)}\in\{0,1\}\) 
indicate whether position \(t\) of the $i$-th sample receives supervision 
and \(s_i \in \{0,1\}\) whether sample \(i\) is selected for distillation. The objective of selective KD (SKD) can be written as:
\begin{align}
\label{eq:skd}
\ell_{\text{SKD}}^{(i)}(t)
&=\lambda\,\mathrm{KL}_{\mathcal{C}_t^{(i)}}\!\left(p_t \,\|\, q_t\right)\\
&\quad + (1-\lambda)\,\mathrm{CE}(y_t, q_t)\nonumber \\
\mathcal{L}_{\text{SKD}}^{(i)}
&=\frac{1}{\sum_{t=1}^{L_i-1} m_t^{(i)}}
\sum_{t=1}^{L_i-1} m_t^{(i)}\,\ell_{\text{SKD}}^{(i)}(t) \\
\mathcal{L}_{\text{SKD}}
&=\frac{1}{\sum_{i=1}^{|\mathcal{D}|} s_i}\sum_{i=1}^{|\mathcal{D}|} s_i\,
\mathcal{L}_{\text{SKD}}^{(i)}
\end{align}
The primary question is  \textit{how to choose classes, positions, and samples} for distillation, namely, how to construct \(\mathcal{C}_t\), \(m_t^{(i)}\), and \(s_i\).

\paragraph{Key Choices for Selective Distillation}
We decompose selective KD  into five orthogonal design choices that determine how teacher information is transferred to the student:
\begin{enumerate}
[leftmargin=*,topsep=0pt,itemsep=3pt,parsep=0pt]
    \item \textit{Alignment criterion}: the objective used for teacher--student alignment, e.g., KL-based or Decoupled KD.

    \item \textit{Position axis}: which token positions receive distillation, i.e., how to choose $m_t^{(i)}$.
    We study this axis via
    (i) the position-importance metric $u(t)$, which quantifies the importance of each position \(t\), and
    (ii) the position-selection policy, namely, a rule that maps the scores \(u(t)\) for a given sample to the values \(m_t^{(i)}\).

    \item \textit{Class axis}: how the teacher distribution over the vocabulary is sparsified at each position, choosing $\mathcal{C}_t^{(i)}$.

    \item \textit{Sample axis}: which training examples are distilled, i.e., how to choose $s_i$.

    \item \textit{Feature axis} (\textit{not explored here}): which  teacher and student representations are being aligned. Beyond next-token distributions, KD can align intermediate features, such as hidden states or attention maps \citep{romero2015fitnets,jiao2020tinybert}. While selection can be applied on this axis as well (e.g., choosing layers or heads), we leave this direction for future work.
\end{enumerate}

Table~\ref{tab:method-overview} summarizes prior methods for selective KD in terms of our framework.
Notably, we observe that no prior work has exploited selection across more than a single axis.
Moreover, student entropy as a distillation signal is underexplored, despite evidence for its effectiveness in training \citep{wang2025highentropy}.
We tackle these gaps next.

\section{Student Entropy Guided Selective KD}

Given the gaps in prior work, we introduce a selective KD method that leverages student entropy as a position-importance signal and employ selection across axes.

\paragraph{Student Entropy-based Position Selection (\selector{})}
We use student entropy to score position importance, i.e., \(u(t)=H(q_t)\). Given a sample $i$ of length \(L\), \selector{} selects the top-$k\%$ most uncertain positions for distillation:
\begin{equation}
m_t^{(i)} = \mathbb{I}\!\left[u(t)\ge \tau\right],
\end{equation}
where \(\tau\) is chosen such that exactly \(\lceil k(L-1)\rceil\) positions satisfy \(m_t^{(i)}{=}1\).
We additionally use a per-sequence normalization in the loss to ensure a fixed supervision budget.

\paragraph{Cross-Axis Selection}
In addition to position selection, we extend \selector{} to operate across the three axes of classes, positions, and samples.
Specifically, we apply class selection via per-token class sampling (\(\mathcal{C}_t^{(i)}\)) using RS-KD, and sample selection via top-$\ell\%$ ranking by average student entropy computed in a single forward-pass preprocessing step using a frozen student, then distilling on the top-$\ell\%$ samples.
We call this variant \triaxselector{}.
These extensions are orthogonal to position selection and enable a unified multi-axis KD that improves accuracy, efficiency, and storage cost.

\paragraph{Selective LM Head and Chunked Entropy Computation}
\label{sec:selective-lm-head}
Selective KD enables two simple, selection-aware optimizations for reducing the logit-related memory footprint.
Let $B$ denote the batch size, $L$ the sequence length, and $V=|\mathcal{V}|$ the vocabulary size.

First, \emph{chunked entropy computation} computes per-position entropy without materializing the full $[B,L,V]$ sized logits tensor: the student hidden states are projected through the LM head in small chunks with gradients disabled, reduced to $O(BL)$ entropy scalars, and discarded.

Second, a \emph{selective LM head} computes logits only at the positions across the batch $N_{\mathrm{select}}$: teacher logits shrink from $[B,L,V]$ to $[N_{\mathrm{select}},V]$, and for the student it computes logits \emph{with gradients enabled} only at selected positions, so the KL loss backpropagates through $N_{\mathrm{select}}$ positions rather than all $BL$, reducing both forward and backward cost.

\section{Experiments}
\label{sec:experiments}

We conduct comprehensive experiments to assess selective KD methods along the axes defined in \S\ref{sec:framework}.
Notably, the design space is large even under conservative choices, spanning position-importance metrics, position-selection policies, and class/sample selection, which yields hundreds of configurations and makes exhaustive evaluation infeasible.
We therefore use a controlled evaluation protocol in which we fix all but one axis at a time. This allows us to isolate the effect of each design choice.

\paragraph{Methods}
We evaluate all the position importance metrics and selection policies in Table~\ref{tab:method-overview}. Except for GLS, position selection is always normalized per sequence length.
Unless stated otherwise, all models are trained with the same hyperparameters described in \S\ref{sec:hparams}.
Below are additional details on the position selection policies:
\begin{itemize}
[leftmargin=*,topsep=0pt,itemsep=3pt,parsep=0pt]
    
    \item \textbf{GLS}: Maintains a queue of recent entropy values and sets \(\tau\) to the empirical \((100{-}k)\)-th percentile of this global distribution to stabilize top-$k$ selection across batches.

    \item \textbf{Pos RS-KD}: A stochastic position-selection policy inspired by RS-KD, sampling positions with probability \(q(t)=\frac{H(q_t)}{\sum_{j} H(q_j)}\). While treated here as a selection policy, repeated sampling induces implicit loss reweighting, yielding an unbiased estimator of weighted KD (see \S\ref{sec:appendix}). 
    
    \item \textbf{Pos RS-KD$^*$}: Importance-corrected variant: after sampling positions with probability \(q(t)\), each sampled position loss is reweighted by \(1/q(t)\), yielding an unbiased estimator of Full KD.

    \item \textbf{Curriculum}: 
    A curriculum-style position-selection method with a fixed budget of $k{=}20\%$ positions per sequence, gradually shifting supervision from low to high-student-entropy tokens over training.

\end{itemize}

\paragraph{Baselines and Ablations}
We compare against the following baselines and component ablations:
\begin{itemize}
[leftmargin=*,topsep=0pt,itemsep=3pt,parsep=0pt]
    \item Off-the-shelf student without distillation, and the teacher as an upper bound.
    \item \textbf{Full KD}: Supervised KD applied densely over all classes, positions, and samples.
    \item \textbf{AT-KD}: As a representative uncertainty-guided position-weighting method.
    \item \textbf{RS-KD}: Class-axis selective distillation using importance sampling over teacher logits.
    \item \textbf{RandomPos \(k\%\)}: Random position selection supervising a fixed fraction \(k\%\) of positions per sample.
    \item \textbf{TopSmp \(\ell\%\)}: Student entropy-based sample selection. This is an ablation of \triaxselector{} that removes class sampling (RS-KD) and position selection.
    \item \textbf{RandomSmp \(\ell\%\)}: Random sample selection supervising a fixed fraction \(\ell\%\) of training samples.
\end{itemize}

We separate global configuration selection from final evaluations.
All methods share the same distillation setup: KD hyperparameters (e.g. temperature \(T=1.0\) and loss weighting \(\lambda=1.0\), yielding a KL-only distillation objective) are selected once on validation data and then fixed, with no method-specific tuning (see \S\ref{sec:additional-results} for details).
Supervision budgets for top-$k\%$ position selection and top-$\ell\%$ sample selection are chosen via a search on validation splits using a single run per setting (see results in \S\ref{sec:appendix-val}), and then fixed for all main comparisons.

\paragraph{Evaluation}
We consider two distillation setups:

(1) \emph{General-purpose distillation} on a large pretraining-style corpus. We train all models on 80 million tokens from FineWeb-Edu~\citep{penedo2024fineweb} and evaluate them in a zero-shot setting.
Documents are packed into sequences of up to 512 tokens.
We measure performance on HellaSwag~\citep{zellers2019hellaswag}, PIQA~\citep{bisk2019piqa}, and Arc-E~\citep{clark2018arc} (multiple-choice reasoning); GSM8K~\citep{cobbe2021gsm8k} (math reasoning); and LAMBADA~\citep{paperno2016lambada} (long-range prediction), reporting average accuracy.
For LAMBADA, we additionally report perplexity and expected calibration error (ECE;~\citealp{guo2017calibration}).
We also evaluate instruction-following on IFEval~\citep{zhou2023ifeval} reporting Pass@1 according to the official verifier\footnote{\url{https://github.com/google-research/google-research/tree/master/ifeval}}.
All results are averaged over three random seeds, with standard deviations reported in \S\ref{sec:std}.

(2) \emph{Task-specific distillation} on a downstream reasoning task. We 
apply KD directly on the GSM8K training set~\citep{cobbe2021gsm8k} and report exact-match accuracy on the GSM8K test set.
In addition to standard off-policy distillation, we evaluate on-policy distillation~\citep{agarwal2024onpolicy}.
We exclude \triaxselector{} from this evaluation, as class-level sampling relies on an offline teacher cache that is incompatible with dynamic student text generation.

\paragraph{Models}
We follow prior work \citep{chen2025distillingessenceefficientreasoning, lu2025onpolicydistillation} and use Qwen3-1.7B as a student and Qwen3-8B as a teacher \citep{yang2025qwen3technicalreport}.

\begin{table}[t]
\centering
\caption{Evaluation results of various \textbf{position-importance metrics with Top-20\% hard selection.}
The best student method is in \textbf{bold},  the second best is \underline{underlined}, and \textbf{\textit{bold italic}} denotes the teacher (upper bound).
Standard deviation values are in \S\ref{sec:std}.}
\footnotesize
\resizebox{\linewidth}{!}{%
\begin{tabular}{lcccc}
\toprule
\textbf{Method} 
& \textbf{Acc. $\uparrow$}
& \textbf{IFEval $\uparrow$}
& \textbf{PPL $\downarrow$}
& \textbf{ECE $\downarrow$} \\
\midrule
Qwen3 1.7B & 61.9 & 19.4 & 12.2 & 30.5 \\
Qwen3 8B & \textbf{\textit{73.8}} & \textbf{\textit{28.9}} & \textbf{\textit{4.6}} & \textbf{\textit{23.5}} \\
Full KD & 64.4 & 20.5 & 7.3 & 27.3 \\
RandomPos 20\% & 64.2 & 20.2 & 7.7 & 27.2 \\
AT-KD & 63.8 & 19.8 & 7.3 & \textbf{26.7} \\
\cmidrule(lr){1-5}
\multicolumn{5}{l}{\textit{Position selection policy: Top 20\%}} \vspace{3px} \\
Student entropy (\selector{}) & \underline{64.8} & \underline{21.4} & 6.9 & 27.6 \\
Teacher entropy & 63.2 & 20.5 & 9.4 & 27.3 \\
Student CE & 63.8 & 20.4 & 8.1 & 27.8 \\
Teacher CE & 63.4 & 19.4 & 9.3 & 27.8 \\
KL & 64.5 & 21.0 & 7.2 & \textbf{26.7} \\
Reverse KL & 64.7 & 20.7 & 6.8 & \underline{27.0} \\
CE ratio & 64.6 & \textbf{22.5} & \textbf{6.5} & 27.7 \\
CE ratio + student entropy & 64.6 & \underline{21.4} & \underline{6.7} & 27.5 \\
Student entropy + KL & \textbf{65.1} & 20.9 & 6.8 & 26.9 \\
\bottomrule
\end{tabular}
}
\label{tab:uncertainty-analysis-test}
\end{table}

\begin{table}[t]
\setlength\tabcolsep{4pt}
\centering
\caption{Evaluation results of \textbf{position-selection policies, applied with student-entropy} as position-importance metric and distillation budget of 20\%, against baselines.
}
\footnotesize
\begin{tabular}{lcccc}
\toprule
\textbf{Method} & \textbf{Acc. $\uparrow$} & \textbf{IFEval $\uparrow$} & \textbf{PPL $\downarrow$} & \textbf{ECE $\downarrow$} \\
\midrule
Qwen3 1.7B & 61.9 & 19.4 & 12.2 & 30.5 \\
Qwen3 8B & \textbf{\textit{73.8}} & \textbf{\textit{28.9}} & \textbf{\textit{4.6}} & \textbf{\textit{23.5}} \\
Full KD & 64.4 & 20.5 & \underline{7.3} & 27.3 \\
RandomPos 20\% & 64.2 & 20.2 & 7.7 & 27.2 \\
AT-KD & 63.8 & 19.8 & \underline{7.3} & \textbf{26.7} \\
\cmidrule(lr){1-5}
\multicolumn{5}{l}{\textit{Position importance metric: Student entropy, \(k=20\%\)}\vspace{3px}} \\
Top 20\% (\selector{}) & \textbf{64.8} & \textbf{21.4} & \textbf{6.9} & 27.6 \\
Top 20\% GLS 30K & 64.5 & \underline{20.7} & 7.5 & 27.6 \\
Curriculum 20\% & \underline{64.6} & \underline{20.7} & \textbf{6.9} & 27.7 \\
Pos RS-KD$^*$ 20\% & 63.6 & 20.6 & 8.3 & 27.6 \\
Pos RS-KD 20\% & 63.0 & 20.1 & 9.9 & \underline{27.0} \\
\bottomrule
\end{tabular}
\label{tab:student-entropy-variants}
\end{table}

\section{Results}

\paragraph{Comparing Position-Importance Metrics}
We begin by fixing the selection policy and budget to top-\(20\%\) and comparing the position-importance metrics.
Table~\ref{tab:uncertainty-analysis-test} presents the results, showing that student entropy based signals and teacher--student discrepancy metrics (CE ratio, KL and reverse KL) most reliably identify informative positions: Top-20\% student entropy achieves strong performance (\(64.8\) accuracy, \(6.9\) perplexity), beating Full KD, and RandomPos while top-20\% KL/reverse-KL/CE-ratio remain competitive (\(64.5\text{--}64.7\) accuracy, with best perplexity at \(6.5\)).
In contrast, ranking by teacher entropy/CE underperforms in both accuracy and perplexity.
Notably, calibration differences are small; although AT-KD, KL, and reverse KL achieve the best ECE, the gaps are limited, suggesting that gains mainly stem from better supervision allocation rather than changes in confidence behavior.

\paragraph{Comparing Position Selection Policies}
We compare position-selection policies at a fixed importance metric and budget.
As shown in Table~\ref{tab:student-entropy-variants}, Top-20\% selection by student entropy (\selector{}) yields the strongest overall performance, improving accuracy (\(64.4 \rightarrow 64.8\)), perplexity (\(7.3\rightarrow 6.9\)), and instruction-following (\(20.5\rightarrow 21.4\)).
It outperforms Full KD, random selection, GLS, curriculum scheduling, and AT-KD in accuracy and IFEval, though AT-KD achieves the best calibration, followed by Pos RS-KD and only then \selector{}.
Pos RS-KD and Pos RS-KD$^{*}$ underperform Top-$k\%$, suggesting that naive entropy-proportional sampling can be suboptimal without additional smoothing or coverage constraints (see \S\ref{sec:pos-rs-failure-mode}).
Overall, student-entropy-guided selection is the most reliable position-selection policy at \(k{=}20\%\), supporting the view that dense supervision is suboptimal.

\begin{figure}[t]
    \centering
    \includegraphics[width=0.9\linewidth]{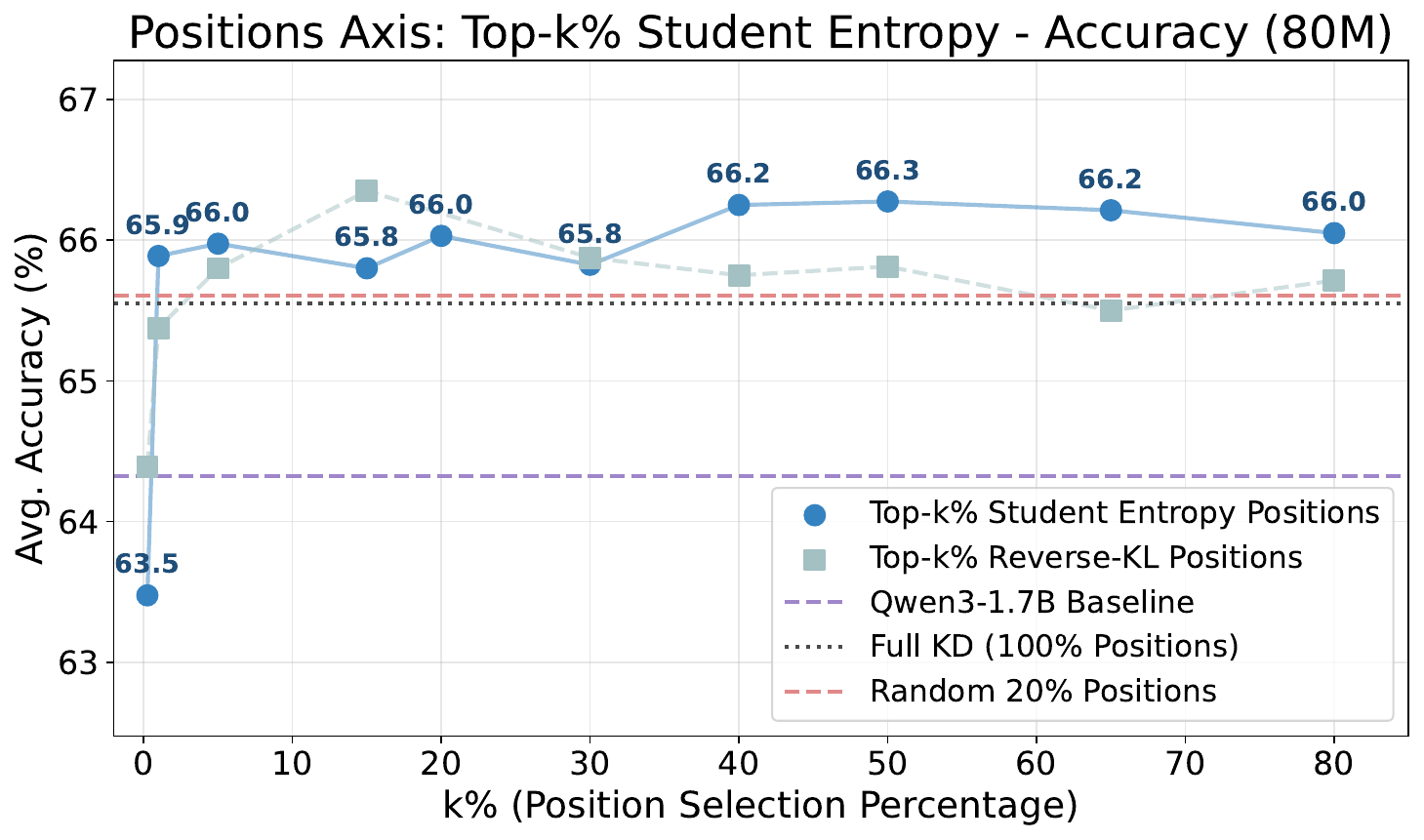}
    \caption{\textbf{Position-axis budget sweep.} Average validation accuracy after distilling on 80M FineWeb-Edu tokens as a function of the supervised position budget \(k\%\). We compare Top-\(k\%\) student-entropy (\selector{}) and Top-\(k\%\) reverse-KL, with Full KD and RandomPos as reference. The teacher accuracy is 77.0.}
    \label{fig:supervised_k_search}
\end{figure}

\begin{figure}[t]
    \centering
    \includegraphics[width=0.9\linewidth]{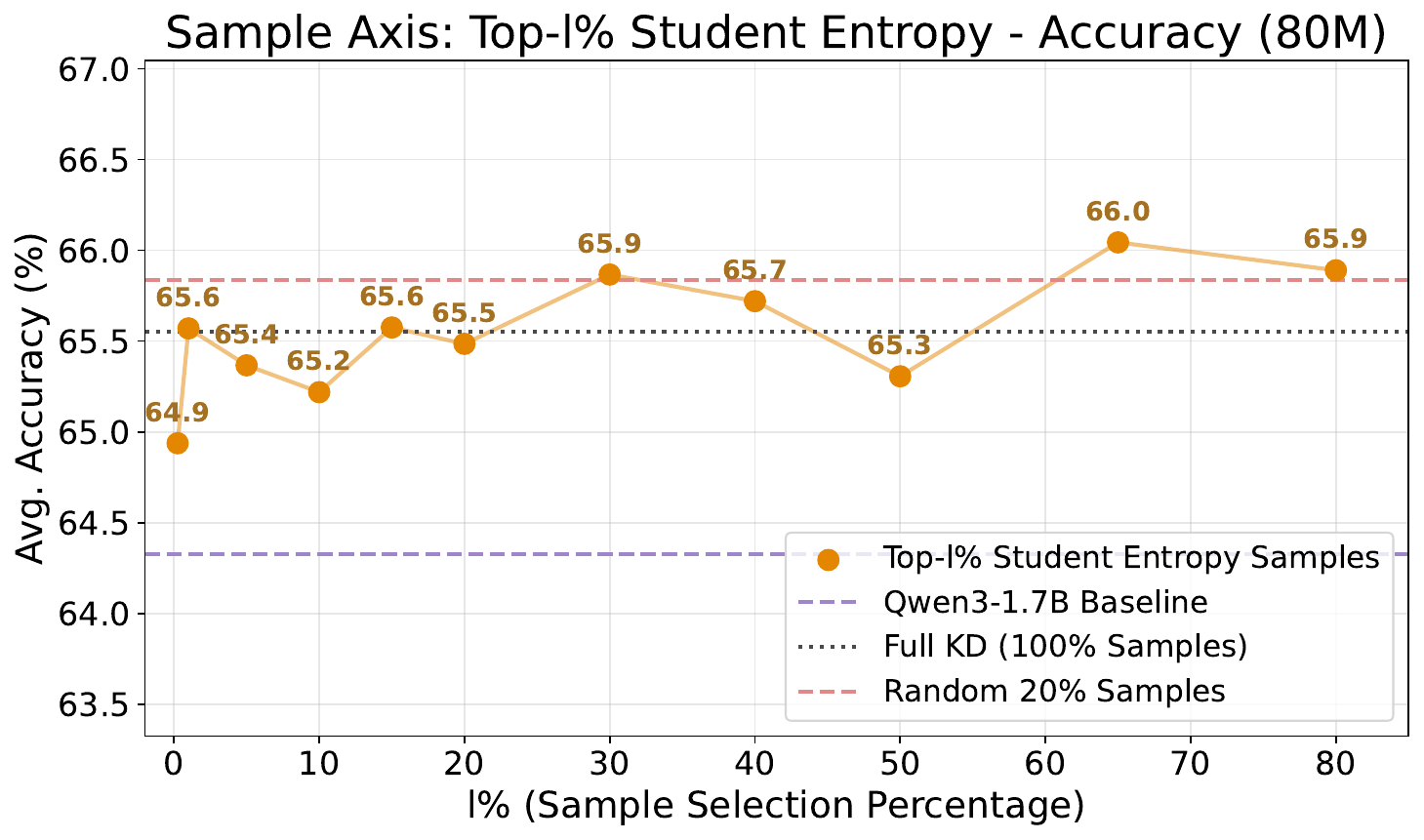}
    \caption{\textbf{Sample-axis budget sweep.} Average validation accuracy after distilling on 80M FineWeb-Edu tokens as a function of the sample-selection budget \(\ell\%\). Only the top-\(\ell\%\) samples ranked by average student entropy are distilled; Full KD and RandomSmp are shown for reference. The teacher accuracy is 77.0.}
    \label{fig:samples_l_search}
\end{figure}

\paragraph{The Effect of Distillation Budget on Performance}
Fig.~\ref{fig:supervised_k_search} and~\ref{fig:samples_l_search} report  the average accuracy on the validation sets (ArcEasy, GSM8K, HellaSwag and PIQA), averaged over multiple runs. 
We show the performance of \selector{} and reverse-KL, as a representative student--teacher discrepancy metric, across varying selection budgets.
The best performance for both methods is obtained for \(k{=}20\%\) (Fig.~\ref{fig:supervised_k_search}), consistent with recent findings that roughly 20\% of high-entropy tokens disproportionately drive learning \citep{wang2025highentropy}.
Both methods are robust across a wide range of \(k\) values, with a shallow optimum at intermediate budgets; notably, supervising as little as \(\sim\!1\%\) of positions already matches or exceeds Full KD, while extremely small budgets (e.g., \(\sim\!0.25\%\)) remain closer to the undistilled baseline.
We use \(k{=}20\%\) in subsequent experiments as a strong accuracy--compute trade-off (near the plateau) and to stay consistent with prior ``small-fraction'' findings \citep{wang2025highentropy}.
We also vary the \emph{sample-axis} budget by distilling only the top-$\ell\%$ samples ranked by average student entropy (Fig.~\ref{fig:samples_l_search}).
Accuracy changes little with $\ell$, while compute scales roughly linearly, so we use $\ell{=}20\%$ in multi-axis experiments.

\begin{table}[t]
\setlength\tabcolsep{4pt}
\caption{\textbf{Multi-axis selective KD},
comparing \triaxselector{} against baselines and mixes of position selection (\selector{}), class sampling (RS-KD), and sample selection (TopSmp) on general-purpose distillation (test split, 80M tokens).
We report average accuracy across benchmarks (Acc.), instruction-following performance (IFEval), LAMBADA perplexity (PPL), and expected calibration error (ECE).
Standard deviations over three seeds are in \S\ref{sec:std}.}
\centering
\footnotesize
\resizebox{\linewidth}{!}{%
\begin{tabular}{lcccc}
\toprule
\textbf{Method} & \textbf{Acc. $\uparrow$} & \textbf{IFEval $\uparrow$} & \textbf{PPL $\downarrow$} & \textbf{ECE $\downarrow$} \\
\midrule
Qwen3 1.7B & 61.9 & 19.4 & 12.2 & 30.5 \\
Qwen3 8B & 73.8 & 29.0 & 4.6 & 23.5 \\
AT-KD & 63.8 & 20.7 & 7.4 & \textbf{26.7} \\
Full KD & 64.4 & 20.5 & 7.3 & 27.3 \\
RandomPos 20\% & 64.1 & 19.8 & 7.6 & \underline{27.1} \\
RandomSmp 20\% & 64.0 & 20.6 & 8.2 & 27.5 \\
\cmidrule(lr){1-5}
\selector{} & \textbf{64.8} & \underline{21.4} & \textbf{6.9} & 27.6 \\
RS-KD & \underline{64.7} & 20.9 & 7.4 & 27.3 \\
TopSmp 20\% & 64.2 & 20.8 & 7.4 & 27.8 \\
TopSmp 20\% + RS-KD & 64.1 & 20.9 & 7.4 & 27.7 \\
\selector{} + TopSmp 20\% & 64.6 & \textbf{22.0} & 6.9 & 28.0 \\
\triaxselector{} & 64.4 & 20.7 & \underline{7.3} & 27.9 \\
\bottomrule
\end{tabular}
}
\label{tab:rskd-samples-comparison}
\end{table}

\paragraph{Selection Across Positions, Classes, and Samples}
Table~\ref{tab:rskd-samples-comparison} compares selective KD across the position, class, and sample axes.
Position selection is the dominant performance contributor; our student-entropy \selector{} improves average accuracy from \(64.4\) (Full KD) to \(64.8\), improves instruction-following (\(21.4\) vs.\ \(20.5\)) and reduces PPL (\(6.9\) vs.\ \(7.3\)), with a modest ECE increase (\(27.6\) vs.\ \(27.3\)).
RS-KD improves accuracy and preserves calibration, while TopSmp remains close overall to Full KD but degrades calibration.
Combining all axes, \triaxselector{} achieves competitive performance (\(64.4\) accuracy, \(20.7\) IFEval, \(7.3\) PPL) and slightly worse calibration, while substantially reducing runtime, memory, and storage (see \S\ref{sec:efficiency}).

\paragraph{General-Purpose vs. Task-Specific Distillation}
Table~\ref{tab:gsm8k-finetune-comparison} reports task-specific distillation results on GSM8K, which differ qualitatively from general-purpose distillation on FineWeb-Edu.
In the off-policy regime, Full KD achieves the best GSM8K accuracy (71.6), while entropy-based Top-$20\%$ position selection degrades performance (69.5).
Our strongest method, \selector{} + TopSmp, remains close to Full KD (70.9) despite substantially reduced supervision.
In the on-policy regime, \selector{} + TopSmp attains the highest GSM8K accuracy (71.2), outperforming Full KD (70.6), while average accuracy differences remain small.

In the on-policy setting, combining entropy-guided position selection with sample filtering yields the strongest results.
However, in the off-policy regime, and unlike general-purpose distillation, entropy-guided position selection alone does not consistently outperform Full KD on GSM8K.
Instead, our methods remain close to Full KD after a single epoch despite using substantially less supervision.
We attribute this in part to GSM8K's limited size, which may constrain the benefits of selective distillation and allow them to emerge more clearly with larger datasets or multi-epoch training. We leave this hypothesis for future work.

\begin{table}[t]
\centering
\setlength{\tabcolsep}{4pt}
\caption{\textbf{Results for task-specific distillation on GSM8K.}
We compare off-policy and on-policy KD methods, reporting GSM8K exact-match accuracy, average evaluation suite accuracy, and LAMBADA OAI perplexity. For on-policy distillation, we used the reverse-KL alignment criterion.
Standard deviations are in \S\ref{sec:std}.}
\small
\resizebox{\linewidth}{!}{%
\begin{tabular}{clcccc}
\toprule
& \textbf{Method} & \textbf{GSM8K Acc. $\uparrow$} & \textbf{Acc. $\uparrow$} & \textbf{PPL $\downarrow$} \\
\midrule
& Qwen3 1.7B & 68.2 & 61.9 & 12.2  \\
& Qwen3 8B & \textit{\textbf{87.8}} & \textit{\textbf{73.8}} & \textit{\textbf{4.6}} \\
\midrule
\multirow{8}{*}{\rotatebox{90}{\textit{Off Policy Distill.}}}
& Full KD & \textbf{71.6} & \textbf{64.5} & \textbf{7.8} \\
& RandomPos 20\% & 70.2 & \underline{64.0} & \underline{8.0} \\
\cmidrule(lr){2-5}
& \selector{} & 69.5 & 63.9 & \underline{8.0} \\
& Pos RS-KD 20\% & 70.5 & 63.5 & 8.9 \\
& Pos RS-KD$^*$ 20\% & 69.1 & 63.3 & 9.2 \\
& TopSmp 20\% & 69.0 & 63.6 & 8.6 \\
& \selector{} + TopSmp 20\% & \underline{70.9} & \underline{64.0} & 8.6 \\
& \triaxselector{} & 70.2 & 63.9 & 8.6 \\
\midrule
\multirow{7}{*}{\rotatebox{90}{\textit{On Policy Distill.}}}
& Full KD & \underline{70.6} & \textbf{63.7} & \underline{10.0} \\
& RandomPos 20\% & 69.3 & 63.3 & \underline{10.0} \\
\cmidrule(lr){2-5}
& \selector{} & 70.0 & \textbf{63.7} & \textbf{9.5} \\
& Pos RS-KD 20\% & 70.5 & 63.2 & 10.5 \\
& Pos RS-KD$^*$ 20\% & 69.7 & 63.3 & \underline{10.0} \\
& TopSmp 20\% & 70.4 & \textbf{63.7} & 10.1 \\
& \selector{} + TopSmp 20\% & \textbf{71.2} & \underline{63.4} & 10.4 \\
\bottomrule
\end{tabular}
}
\label{tab:gsm8k-finetune-comparison}
\end{table}

\section{Distillation Efficiency}
\label{sec:efficiency}

A major motivation for selective KD is reducing computational costs.
We therefore analyze distillation efficiency in terms of \emph{offline storage} for teacher supervision and \emph{runtime compute} during distillation.
We show that while position selection primarily improves accuracy, sample-level selection yields prominent efficiency gains, and class-level sampling enables orders-of-magnitude reductions in storage.

\subsection{Storage Efficiency}

We follow the formulation of \citet{anshumann2025sparse}, focusing on savings from class- and sample-selection.
Position selection is excluded since it would require storing dynamic uncertainty masks (see \S\ref{sec:offline_cache}).

Storage is measured in bytes per token and reported in decimal terabytes (TB) for a dataset of $N{=}100$B tokens.
Storing a single sampled teacher class requires 24 bits (3 bytes): 17 bits for the vocabulary index and 7 bits for a quantized probability, so caching \(U = |\mathcal{C}_t|\) sampled classes costs $3U$ bytes per position.

Table~\ref{tab:storage} summarizes the storage footprint for Full KD, RS-KD, and \triaxselector{}.
As a baseline, we add vanilla CE training without teacher logits. Unlike \citet{anshumann2025sparse} who used $U{=}12$, we use $U{=}64$ for improved stability, yielding
\(64 \times 3 = 192~\text{bytes/position}.\)
Caching full teacher logits over a vocabulary of size $|\mathcal{V}|{=}100{,}000$ requires $200$~kB per position in float16, making RS-KD with $U{=}64$ roughly $10^3$--$2{\times}10^3$ times more storage-efficient, or $19.2$~TB for $N{=}100$B tokens.
With sample selection, we distill only on the top-$\ell\%$ samples ranked by average student entropy from a single forward pass of a frozen student before distillation. This reduces storage linearly with $\ell$.
For $\ell{=}0.2$, this yields: ${\ell \cdot U \cdot 3 = 38.4~\text{bytes/position}}$, or $3.84$~TB in total.

Overall, RS-KD reduces storage from \(10{,}000\)~TB to \(19.2\)~TB (\(99.8\%\), \(\sim520\times\)) and \triaxselector{} further reduces this to \(3.84\)~TB \((99.96\%\) vs. Full KD and \(80\%\) vs. RS-KD).
Sample indices are also cached but incur negligible storage.

\begin{table}[t]
	\centering
    \caption{\textbf{Offline cache footprint} in terabytes (TB) for $N{=}100$B training tokens and vocabulary size $|\mathcal{V}|{=}100{,}000$.
	RS-KD uses importance sampling over classes; \triaxselector{} further reduces storage via sample-level selection with $\ell{=}20\%$.
    }
    \footnotesize
    \begin{tabular}{lrrrr}
        \toprule
        Method & Classes & TB ($U{=}12$) & TB ($U{=}64$) \\
        \midrule
        Full KD & $|\mathcal{V}|=100K$ & 10{,}000.0 & 10{,}000.0 \\
        RS-KD & $U$ & 3.6 & 19.2 \\
        \triaxselector{} & $U \times 0.2$ & 0.72 & 3.84 \\
        Vanilla CE & 1 & 0.3 & 0.3 \\
        \bottomrule
    \end{tabular}%
	\label{tab:storage}
\end{table}

\begin{table}[t]
\centering
\caption{\textbf{Runtimes and test accuracy} for sample-selection methods (80M tokens, top-20\%, single runs) on GeForce RTX 3090.}
\small
\resizebox{\linewidth}{!}{%
\begin{tabular}{l l l r}
\toprule
\textbf{Method} & \textbf{Sample Selection} & \textbf{Total Wall Time} & \textbf{Acc.} \\
\midrule
Full KD (100\% positions) & 0h00m & 22h52m & \textbf{64.6} \\
RandomPos 20\% & 0h00m & 18h38m & 64.1 \\
\cmidrule(lr){1-4}
TopSmp CE ratio & 8h50m & 13h36m & 64.3 \\
TopSmp KL & 9h37m & 14h42m & 64.2 \\
TopSmp student entropy (ours) & \textbf{2h01m} & \textbf{7h01m} & 64.2 \\
\cmidrule{1-4}
\quad \triaxselector{} (cache construction) & 2h11m & 8h46m & 64.4 \\
\quad \triaxselector{} (reuse offline cache) & 0h00m & 3h58m & 64.4 \\
\bottomrule
\end{tabular}
}
\label{tab:skipping-indices-runtime}
\end{table}

\subsection{Runtime Efficiency}

\paragraph{Runtime Speedups}
\triaxselector{} achieves substantial efficiency gains through sample selection, which directly reduces the number of sequences requiring teacher supervision.
As shown in Table~\ref{tab:skipping-indices-runtime}, this leads to a pronounced reduction in total wall-clock time.
Sample selection incurs an upfront scoring cost: teacher--student metrics require full passes (8h50m for CE ratio, 9h37m for KL), while student-only entropy is cheaper (2h01m).
Reusing an offline cache of selected indices removes this step, reducing \triaxselector{} runtime to 3h58m (Table~\ref{tab:skipping-indices-runtime}).
For a training set of $N$ samples of average length $L$, Full KD supervises $\mathcal{O}(NL)$ positions, reduced to $\mathcal{O}(\ell NL)$ with top-$\ell\%$ sample selection.
Therefore, sample selection provides the main efficiency gains while position selection adds further speedups (up to $\sim$30\% with a selective LM head and chunked entropy; see~\S\ref{sec:memory_utilization}).

\paragraph{Memory Savings of \selector{}}
Position selection primarily reallocates the KD signal \emph{within} a sequence.
While the student and teacher still process the full context, selection reduces the number of positions that require logit computation and gradient-carrying KD terms. This enables memory-oriented implementations that substantially reduce the peak logit-related memory footprint.

In our setting ($B{=}2$, $L{=}512$), selective LM heads with chunked entropy at $k{=}20\%$ reduce the sum of per-GPU peak memory allocations by 18.3\% (33.18\,GB $\rightarrow$ 27.10\,GB): student peak drops by 28.1\% (15.88\,GB $\rightarrow$ 11.42\,GB) and teacher peak by 9.4\% (17.30\,GB $\rightarrow$ 15.68\,GB).
The gains come from avoiding full $[B,L,V]$ logit materialization during selection and restricting KD logits/backprop to the $N_{\mathrm{sel}}$ selected positions.
See~\S\ref{sec:memory_utilization} for ablations and memory traces.

\section{Conclusion and Discussion}

We revisit selective knowledge distillation for autoregressive LLMs through a unified framework that disentangles where and how teacher supervision is applied.
Across a systematic study, we find that dense, uniform logit supervision is often unnecessary: for general-purpose distillation, concentrating supervision on a small subset of high-uncertainty positions consistently matches or outperforms Full KD.

Student-entropy-guided Top-20\% selection is the most reliable overall strategy, while curriculum learning, CE-ratio ranking, and teacher--student KL are promising alternatives.
We also show that position selection integrates effectively with class- and sample-level sparsification, yielding favorable accuracy--efficiency trade-offs; in particular, \triaxselector{} enables substantial speedups via sample filtering and offline teacher caching, and can be implemented with reduced peak memory through a selective LM head.

\paragraph{Limitations and Future Work}
The selective KD design space is large; to keep comparisons controlled, we study a single, widely used teacher--student pair and a fixed supervision budget.
Validating the trends across additional model families, scales, and longer contexts is an important next step.
Selective policies may also interact with alternative alignment criteria (e.g., feature-based KD), and the smaller performance degradation we observe in task-specific distillation suggest further optimizations are needed.

\section*{Impact Statement}
This paper aims to advance knowledge distillation for large language models.
We do not identify societal impacts specific to this work beyond the general considerations associated with training and deploying language models.

\bibliography{custom}

@Article{hinton2015distillation,
  title   = {Distilling the Knowledge in a Neural Network},
  author  = {Hinton, Geoffrey and Vinyals, Oriol and Dean, Jeff},
  journal = {arXiv preprint arXiv:1503.02531},
  year    = {2015},
  url     = {https://arxiv.org/abs/1503.02531}
}

@inproceedings{agarwal2024onpolicy,
    title={On-Policy Distillation of Language Models: Learning from Self-Generated Mistakes},
    author={Rishabh Agarwal and Nino Vieillard and Yongchao Zhou and Piotr Stanczyk and Sabela Ramos Garea and Matthieu Geist and Olivier Bachem},
    booktitle={The Twelfth International Conference on Learning Representations},
    year={2024},
    url={https://openreview.net/forum?id=3zKtaqxLhW}
}

@Article{anshumann2025sparse,
  title   = {Sparse Logit Sampling: Accelerating Knowledge Distillation in {LLM}s},
  author  = {Anshumann, A. and Zaidi, Mohd Abbas and Kedia, Akhil and Ahn, Jinwoo and Kwon, Taehwak and Lee, Kangwook and Lee, Haejun and Lee, Joohyung},
  journal = {arXiv preprint arXiv:2503.16870},
  year    = {2025},
  url     = {https://arxiv.org/abs/2503.16870}
}

@Article{wang2021selectivekd,
  title   = {Selective Knowledge Distillation for Neural Machine Translation},
  author  = {Wang, Fusheng and Yan, Jianhao and Meng, Fandong and Zhou, Jie},
  journal = {arXiv preprint arXiv:2105.12967},
  year    = {2021},
  url     = {https://arxiv.org/abs/2105.12967}
}

@Article{feng2024kpod,
  title   = {Keypoint-based Progressive Chain-of-Thought Distillation for {LLM}s},
  author  = {Feng, Kaituo and Li, Changsheng and Zhang, Xiaolu and Zhou, Jun and Yuan, Ye and Wang, Guoren},
  journal = {arXiv preprint arXiv:2405.16064},
  year    = {2024},
  url     = {https://arxiv.org/abs/2405.16064}
}

@inproceedings{guo2017calibration,
  title     = {On Calibration of Modern Neural Networks},
  author    = {Guo, Chuan and Pleiss, Geoff and Sun, Yu and Weinberger, Kilian Q.},
  booktitle = {Proceedings of the 34th International Conference on Machine Learning (ICML)},
  series    = {Proceedings of Machine Learning Research},
  volume    = {70},
  pages     = {1321--1330},
  year      = {2017},
  publisher = {PMLR},
  url       = {https://proceedings.mlr.press/v70/guo17a.html}
}

@misc{raman2023slim,
  title        = {For Distillation, Tokens Are Not All You Need},
  author       = {Raman, Neeraj and Vare, Siddharth and Srinivasan, Apurva and Chandra, Vignesh and Khandelwal, Kush},
  howpublished = {OpenReview},
  year         = {2023},
  url          = {https://openreview.net/pdf?id=2fc5GOPYip}
}

@inproceedings{shum2024first,
  title     = {{FIRST}: Teach A Reliable Large Language Model Through Efficient Trustworthy Distillation},
  author    = {Shum, Kashun and Xu, Minrui and Zhang, Jianshu and Chen, Zixin and Diao, Shizhe and Dong, Hanze and Zhang, Jipeng and Raza, Muhammad Omer},
  booktitle = {Proceedings of the 2024 Conference on Empirical Methods in Natural Language Processing (EMNLP)},
  pages     = {12646--12659},
  year      = {2024},
  publisher = {Association for Computational Linguistics},
  url       = {https://aclanthology.org/2024.emnlp-main.703.pdf}
}

@Article{su2023eakd,
  title   = {{EA}-{KD}: Entropy-based Adaptive Knowledge Distillation},
  author  = {Su, Cheng-Wei and Tseng, Shih-Hsin and Martins, Jo{\~a}o Vitor and Ichimura, Naoyuki and Seiji, Yokou and Chou, Chien-Hsing},
  journal = {arXiv preprint arXiv:2311.13621},
  year    = {2023},
  url     = {https://arxiv.org/abs/2311.13621}
}

@article{xu2020unix,
    title = {Computation-Efficient Knowledge Distillation via Uncertainty-Aware Mixup},
    journal = {Pattern Recognition},
    volume = {138},
    pages = {109338},
    year = {2023},
    author = {Guodong Xu and Ziwei Liu and Chen Change Loy},
    issn = {0031-3203},
    doi = {https://doi.org/10.1016/j.patcog.2023.109338},
    url = {https://www.sciencedirect.com/science/article/pii/S0031320323000390},
    author = {Guodong Xu and Ziwei Liu and Chen Change Loy},
}

@Article{guo2024entropykd,
  title   = {Leveraging Logit Uncertainty for Better Knowledge Distillation},
  author  = {Guo, Zhen and Wang, Dong and He, Qiang and Zhang, Pengzhou},
  journal = {Scientific Reports},
  volume  = {14},
  number  = {31249},
  year    = {2024},
  doi     = {10.1038/s41598-024-82647-6},
  url     = {https://www.nature.com/articles/s41598-024-82647-6}
}

@Article{cobbe2021gsm8k,
  author       = {Karl Cobbe and
                  Vineet Kosaraju and
                  Mohammad Bavarian and
                  Mark Chen and
                  Heewoo Jun and
                  Lukasz Kaiser and
                  Matthias Plappert and
                  Jerry Tworek and
                  Jacob Hilton and
                  Reiichiro Nakano and
                  Christopher Hesse and
                  John Schulman},
  title        = {Training Verifiers to Solve Math Word Problems},
  journal      = {CoRR},
  volume       = {abs/2110.14168},
  year         = {2021},
  url          = {https://arxiv.org/abs/2110.14168},
  eprinttype    = {arXiv},
  eprint       = {2110.14168},
  bibsource    = {dblp computer science bibliography, https://dblp.org}
}

@inproceedings{paperno2016lambada,
  title     = {The LAMBADA dataset: Word prediction requiring a broad discourse context},
  author    = {Paperno, Denis and Kruszewski, Germ{\'a}n and Lazaridou, Angeliki and Pham, Quan and Bernardi, Raffaella and Pezzelle, Sandro and Baroni, Marco and Boleda, Gemma and Fern{\'a}ndez, Raquel},
  booktitle = {Proceedings of the 54th Annual Meeting of the Association for Computational Linguistics (ACL)},
  year      = {2016},
  url       = {https://aclanthology.org/P16-1144}
}

@inproceedings{zellers2019hellaswag,
  title     = {HellaSwag: Can a Machine Really Finish Your Sentence?},
  author    = {Zellers, Rowan and Bisk, Yonatan and Farhadi, Ali and Choi, Yejin},
  booktitle = {Proceedings of the 57th Annual Meeting of the Association for Computational Linguistics (ACL)},
  year      = {2019},
  url       = {https://aclanthology.org/P19-1472}
}

@misc{bisk2019piqa,
  title         = {PIQA: Reasoning about Physical Commonsense in Natural Language},
  author        = {Yonatan Bisk and Rowan Zellers and Ronan Le Bras and Jianfeng Gao and Yejin Choi},
  year          = {2019},
  eprint        = {1911.11641},
  archiveprefix = {arXiv},
  primaryclass  = {cs.CL},
  url           = {https://arxiv.org/abs/1911.11641}
}

@inproceedings{clark2018arc,
  title     = {Think you have Solved Question Answering? Try ARC, the AI2 Reasoning Challenge},
  author    = {Clark, Peter and Cowhey, Isaac and Etzioni, Oren and Khot, Tushar and Sabharwal, Ashish and Schoenick, Carissa and Tafjord, Oyvind},
  booktitle = {Proceedings of the 56th Annual Meeting of the Association for Computational Linguistics (ACL)},
  year      = {2018},
  url       = {https://aclanthology.org/P18-1260}
}

@misc{zhou2023ifeval,
      title={Instruction-Following Evaluation for Large Language Models}, 
      author={Jeffrey Zhou and Tianjian Lu and Swaroop Mishra and Siddhartha Brahma and Sujoy Basu and Yi Luan and Denny Zhou and Le Hou},
      year={2023},
      eprint={2311.07911},
      archivePrefix={arXiv},
      primaryClass={cs.CL},
      url={https://arxiv.org/abs/2311.07911}, 
}

@Article{wang2025highentropy,
  title   = {Beyond the 80/20 Rule: High-Entropy Minority Tokens Drive Effective Reinforcement Learning for LLM Reasoning},
  author  = {Wang, Shenzhi and Yu, Le and Gao, Chang and Zheng, Chujie and Liu, Shixuan and Lu, Rui and Dang, Kai and Chen, Xionghui and Yang, Jianxin and Zhang, Zhenru and Liu, Yuqiong and Yang, An and Zhao, Andrew and Yue, Yang and Song, Shiji and Yu, Bowen and Huang, Gao and Lin, Junyang},
  journal = {arXiv preprint arXiv:2506.01939},
  year    = {2025},
  url     = {https://arxiv.org/abs/2506.01939}
}

@Article{zhong2024atkd,
  title   = {Revisiting Knowledge Distillation for Autoregressive Language Models},
  author  = {Zhong, Qihuang and Ding, Liang and Shen, Li and Liu, Juhua and Du, Bo and Tao, Dacheng},
  journal = {arXiv preprint arXiv:2402.11890},
  year    = {2024},
  url     = {https://arxiv.org/abs/2402.11890}
}

@inproceedings{zhao2022decoupled,
  title     = {Decoupled Knowledge Distillation},
  author    = {Zhao, Borui and Cui, Quan and Song, Renjie and Qiu, Yiyu and Liang, Jiajun},
  booktitle = {Proceedings of the IEEE/CVF Conference on Computer Vision and Pattern Recognition (CVPR)},
  pages     = {11953--11962},
  year      = {2022},
  url       = {https://arxiv.org/abs/2203.08679}
}

@Article{lu2025onpolicydistillation,
  author = {Kevin Lu and Thinking Machines Lab},
  title = {On-Policy Distillation},
  journal = {Thinking Machines Lab: Connectionism},
  year = {2025},
  note = {https://thinkingmachines.ai/blog/on-policy-distillation},
  doi = {10.64434/tml.20251026},
}

@misc{yang2025qwen3technicalreport,
      title={Qwen3 Technical Report}, 
      author={An Yang and Anfeng Li and Baosong Yang and Beichen Zhang and Binyuan Hui and Bo Zheng and Bowen Yu and Chang Gao and Chengen Huang and Chenxu Lv and Chujie Zheng and Dayiheng Liu and Fan Zhou and Fei Huang and Feng Hu and Hao Ge and Haoran Wei and Huan Lin and Jialong Tang and Jian Yang and Jianhong Tu and Jianwei Zhang and Jianxin Yang and Jiaxi Yang and Jing Zhou and Jingren Zhou and Junyang Lin and Kai Dang and Keqin Bao and Kexin Yang and Le Yu and Lianghao Deng and Mei Li and Mingfeng Xue and Mingze Li and Pei Zhang and Peng Wang and Qin Zhu and Rui Men and Ruize Gao and Shixuan Liu and Shuang Luo and Tianhao Li and Tianyi Tang and Wenbiao Yin and Xingzhang Ren and Xinyu Wang and Xinyu Zhang and Xuancheng Ren and Yang Fan and Yang Su and Yichang Zhang and Yinger Zhang and Yu Wan and Yuqiong Liu and Zekun Wang and Zeyu Cui and Zhenru Zhang and Zhipeng Zhou and Zihan Qiu},
      year={2025},
      eprint={2505.09388},
      archivePrefix={arXiv},
      primaryClass={cs.CL},
      url={https://arxiv.org/abs/2505.09388}, 
}

@Article{penedo2024fineweb,
  title   = {The FineWeb Datasets: Decanting the Web for the Finest Text Data at Scale},
  author  = {Penedo, Guilherme and Kydl{\'i}{\v{c}}ek, Hynek and Ben Allal, Loubna and Lozhkov, Anton and Mitchell, Margaret and Raffel, Colin and Von Werra, Leandro and Wolf, Thomas},
  journal = {arXiv preprint arXiv:2406.17557},
  year    = {2024},
  url     = {https://arxiv.org/abs/2406.17557}
}

@misc{huang2025selectkd,
      title={SelecTKD: Selective Token-Weighted Knowledge Distillation for LLMs}, 
      author={Haiduo Huang and Jiangcheng Song and Yadong Zhang and Pengju Ren},
      year={2025},
      eprint={2510.24021},
      archivePrefix={arXiv},
      primaryClass={cs.CL},
      url={https://arxiv.org/abs/2510.24021}, 
}

@misc{xie2025adakd,
  title        = {LLM-Oriented Token-Adaptive Knowledge Distillation},
  author       = {Xurong Xie and Zhucun Xue and Jiafu Wu and Jian Li and Yabiao Wang and Xiaobin Hu and Yong Liu and Jiangning Zhang},
  year         = {2025},
  eprint       = {2510.11615},
  archivePrefix= {arXiv},
  primaryClass = {cs.CL},
  url          = {https://arxiv.org/abs/2510.11615}
}

@misc{chen2025distillingessenceefficientreasoning,
      title={Distilling the Essence: Efficient Reasoning Distillation via Sequence Truncation}, 
      author={Wei-Rui Chen and Vignesh Kothapalli and Ata Fatahibaarzi and Hejian Sang and Shao Tang and Qingquan Song and Zhipeng Wang and Muhammad Abdul-Mageed},
      year={2025},
      eprint={2512.21002},
      archivePrefix={arXiv},
      primaryClass={cs.CL},
      url={https://arxiv.org/abs/2512.21002}, 
}

@inproceedings{he2025dakd,
    title={{DA}-{KD}: Difficulty-Aware Knowledge Distillation for Efficient Large Language Models},
    author={Changyi He and Yifu Ding and Jinyang Guo and Ruihao Gong and Haotong Qin and Xianglong Liu},
    booktitle={Forty-second International Conference on Machine Learning},
    year={2025},
    url={https://openreview.net/forum?id=NCYBdRCpw1}
}

@inproceedings{romero2015fitnets,
  author       = {Adriana Romero and
                  Nicolas Ballas and
                  Samira Ebrahimi Kahou and
                  Antoine Chassang and
                  Carlo Gatta and
                  Yoshua Bengio},
  editor       = {Yoshua Bengio and
                  Yann LeCun},
  title        = {FitNets: Hints for Thin Deep Nets},
  booktitle    = {3rd International Conference on Learning Representations, {ICLR} 2015,
                  San Diego, CA, USA, May 7-9, 2015, Conference Track Proceedings},
  year         = {2015},
  url          = {http://arxiv.org/abs/1412.6550},
  bibsource    = {dblp computer science bibliography, https://dblp.org}
}

@inproceedings{jiao2020tinybert,
    title = "{T}iny{BERT}: Distilling {BERT} for Natural Language Understanding",
    author = "Jiao, Xiaoqi  and
      Yin, Yichun  and
      Shang, Lifeng  and
      Jiang, Xin  and
      Chen, Xiao  and
      Li, Linlin  and
      Wang, Fang  and
      Liu, Qun",
    editor = "Cohn, Trevor  and
      He, Yulan  and
      Liu, Yang",
    booktitle = "Findings of the Association for Computational Linguistics: EMNLP 2020",
    month = nov,
    year = "2020",
    address = "Online",
    publisher = "Association for Computational Linguistics",
    url = "https://aclanthology.org/2020.findings-emnlp.372/",
    doi = "10.18653/v1/2020.findings-emnlp.372",
    pages = "4163--4174",
}
\bibliographystyle{icml2026}

\newpage
\appendix
\onecolumn

\section{Proof: Positional Random Sampling Selection is an Unbiased Estimator of Weighted KD}
\label{sec:appendix}

In this section, we prove that a knowledge distillation using the positional RS selection method matches the weighted KD in expectation.
It is important to note that one can easily transform such a selection method to match full KD in expectation, but we deliberately do not do so, since we aim to match a weighted KD that emphasizes tokens according to their entropy.

Consider a sequence of length $N$ token positions, indexed by $t \in \{1,\dots,N\}$.
Let $L_t$ denote the per-token distillation loss at position $t$,
and let $w(t)$ be a non-negative importance weight assigned to that token.
We sample $K$ token indices $t_k$ i.i.d.\ from the following distribution:

\[
q(t) = \frac{w(t)}{\sum_{j=1}^N w(j)}.
\]

Here, $q(t)$ denotes the sampling distribution over token positions,
$\widehat{\mathcal{L}}$ is the empirical loss estimator,
and $\mathbb{E}[\cdot]$ denotes expectation over the sampling process.

Using this notation, we have
\begin{align*}
{\mathbb{E}}[L_{t_k}]
&= \sum_{t=1}^N q(t)L_t
= \sum_{t=1}^N \frac{w(t)}{\sum_{j=1}^N w(j)} L_t \\
&= \mathcal{L}_{\text{weighted}}.
\end{align*}
Hence, the probability of sampling token \(t\) is proportional to its contribution in the weighted KD objective.
\begin{align*}
{\mathbb{E}}[\widehat{\mathcal{L}}]
= {\mathbb{E}}\!\left[\frac{1}{K}\sum_{k=1}^K L_{t_k}\right]
= \frac{1}{K} \sum_{k=1}^K \mathbb{E}[L_{t_k}] \\
= \frac{1}{K} \cdot K \cdot \mathcal{L}_{\text{weighted}}
= \mathcal{L}_{\text{weighted}}.
\end{align*}

It can also be viewed by denoting \(c_t\) as how many times token \(t\) was sampled:

\begin{align*}
c_t = \sum_{k=1}^K \mathbf{1}_{t_k=t},
\quad
\widehat{\mathcal{L}} = \frac{1}{K}\sum_{t} c_t L_t, \\
{\mathbb{E}}[c_t] = K q(t).
\end{align*}

So,
\begin{align*}
{\mathbb{E}}[\widehat{\mathcal{L}}]
= \frac{1}{K}\sum_t {\mathbb{E}}[c_t] L_t 
= \frac{1}{K}\sum_t K q(t) L_t \\
= \sum_t q(t)L_t.
\end{align*}

Hence, \(\widehat{\mathcal{L}}\) is an unbiased estimator of the \textbf{weighted KD objective}:
\[
\mathcal{L}_{\text{weighted}}
:= \sum_{t=1}^N q(t)\,L_t
= \frac{\sum_t w(t)\,L_t}{\sum_j w(j)}.
\]

\paragraph{Importance-corrected positional random sampling.}
For completeness, we note that positional random sampling can also be made an unbiased estimator of the full (unweighted) KD objective via importance correction.
Specifically, if each sampled loss is reweighted by the inverse sampling probability,
\(
\widehat{\mathcal{L}}_{\text{IC}}
= \frac{1}{KN}\sum_{k=1}^K \frac{L_{t_k}}{q(t_k)},
\)
then
\(
\mathbb{E}[\widehat{\mathcal{L}}_{\text{IC}}]
= \frac{1}{N}\sum_{t=1}^N L_t,
\)
recovering Full KD in expectation.
We referred to this variant as \emph{importance-corrected positional random sampling} and evaluated it separately in our experiments.

\section{Hyperparameters}
\label{sec:hparams}

Tables~\ref{tab:shared-hparams} and~\ref{tab:variant-hparams} list the hyperparameter choices shared across all runs and the settings that differ between distillation variants.

\begin{table}[h]
\centering
\small
\setlength{\tabcolsep}{8pt}
\begin{tabular}{p{0.25\linewidth}p{0.65\linewidth}}
Component & Value \\
\midrule
Teacher model & Qwen/Qwen3-8B (online, no quantization) \\
Student model & Qwen/Qwen3-1.7B \\
Dataset & FineWeb-Edu stream (80M tokens) \\
Sequence length & 512 tokens (\texttt{max\_seq\_len=512}) \\
Epochs & 1 pass over the streamed subset \\
Mini-batch & batch size 2 $\times$ 8 gradient accumulation steps (effective batch 16) \\
Optimizer & bitsandbytes Adam8bit (lr $1\times 10^{-5}$) \\
KD temperature & 1.0 \\
CE mixing weight & $\alpha_{\text{CE}}=1-\lambda=0.0$ (pure KL divergence loss) \\
Offline cache & Enabled with $U=64$ cached classes \\
Seeds & 1337, 1338, 1339 (or 1340, 1341, 1342 for the GSM8K setup) \\
\bottomrule
\end{tabular}
\caption{Shared hyperparameters across all experiments.}
\label{tab:shared-hparams}
\end{table}

\begin{table}[h]
\centering
\small
\setlength{\tabcolsep}{8pt}
\begin{tabular}{p{0.32\linewidth}p{0.6\linewidth}}
Variant & Additional settings \\
\midrule
Full KD & Distills all tokens ($k=100\%$). \\
SE-KD (student entropy top-$k$) & $k=20\%$; selection normalized by sequence length \\
Curriculum Learning & \texttt{SELECTION\_CURRICULUM\_STEPS}$=4000$. \\
Random token selection & $k=20\%$; uniform random token selection, normalized by length. \\
Pos-RS-KD & $k=20\%$; student entropy scoring; \texttt{POS\_RS\_MATCH\_FULL\_KD}$=1$ for corrected variant. \\
\bottomrule
\end{tabular}
\caption{Settings specific to each distillation variant reported in the main tables.}
\label{tab:variant-hparams}
\end{table}

\section{Additional Results}
\label{sec:additional-results}

This section reports auxiliary experiments that motivate the hyperparameter choices used throughout the paper.
We compare temperature settings and cross-entropy mixing weights for the Full KD baseline.
Across these ablations, temperature $T{=}1.0$ mostly outperforms higher temperatures, and the cross-entropy component provides negligible benefit; moreover, including CE would prevent some of our selection-based efficiency optimizations (e.g., restricting gradient-carrying logits to selected positions).
Accordingly, we use $T{=}1.0$ and set $\lambda{=}1$ in Eq.~\ref{eq:full-kd} (pure KL) in all main experiments.

\begin{figure}[t]
    \centering
    \includegraphics[width=0.6\linewidth]{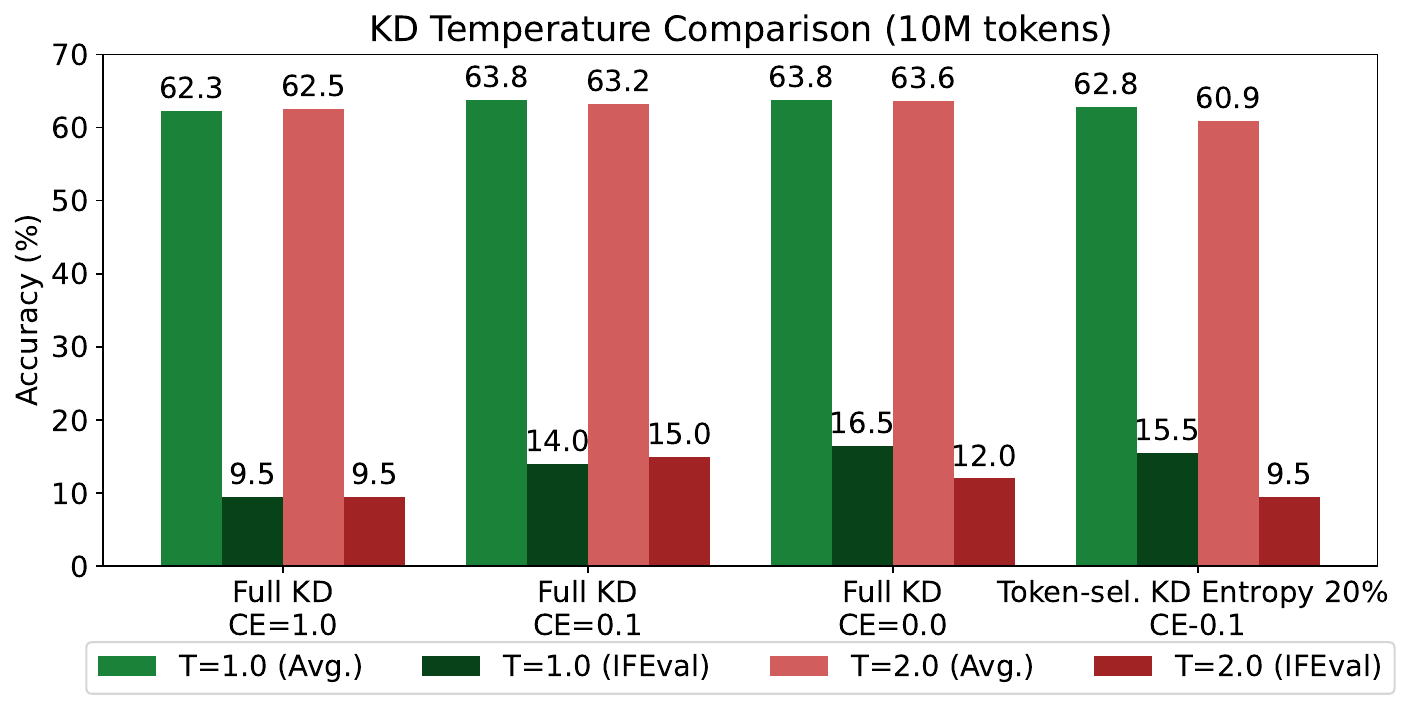}
    \caption{\textbf{Temperature ablation for Full KD.}
    We compare $T{=}2.0$ vs.\ $T{=}1.0$ and report average accuracy over five benchmarks (ArcEasy, GSM8K, HellaSwag, PIQA, and LAMBADA OpenAI).}
    \label{fig:temp-comp}
\end{figure}

\begin{figure*}[t]
    \centering
    \includegraphics[width=\linewidth]{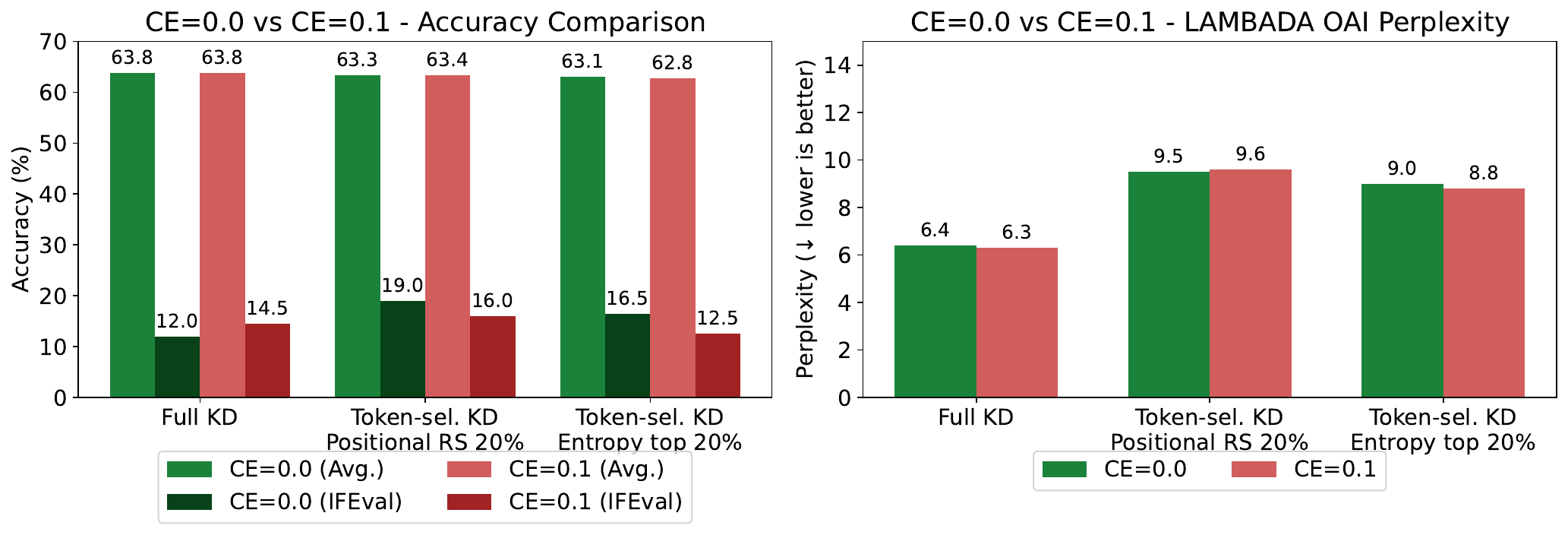}
    \caption{\textbf{Cross-entropy mixing ablation for Full KD.}
    We compare $\alpha_{\mathrm{CE}}=1-\lambda{=}0.1$ vs.\ $\alpha_{\mathrm{CE}}{=}0.0$ and report the same average accuracy metric.
    This study uses a smaller 10M-token run and is included as a sanity check rather than a fully converged comparison.}
    \label{fig:ce-weight-comp}
\end{figure*}

\section{Positional Random Sampling Underperformance}
\label{sec:pos-rs-failure-mode}

\begin{figure*}[t]
    \centering
    \includegraphics[height=\dimexpr\textheight-4\baselineskip\relax]{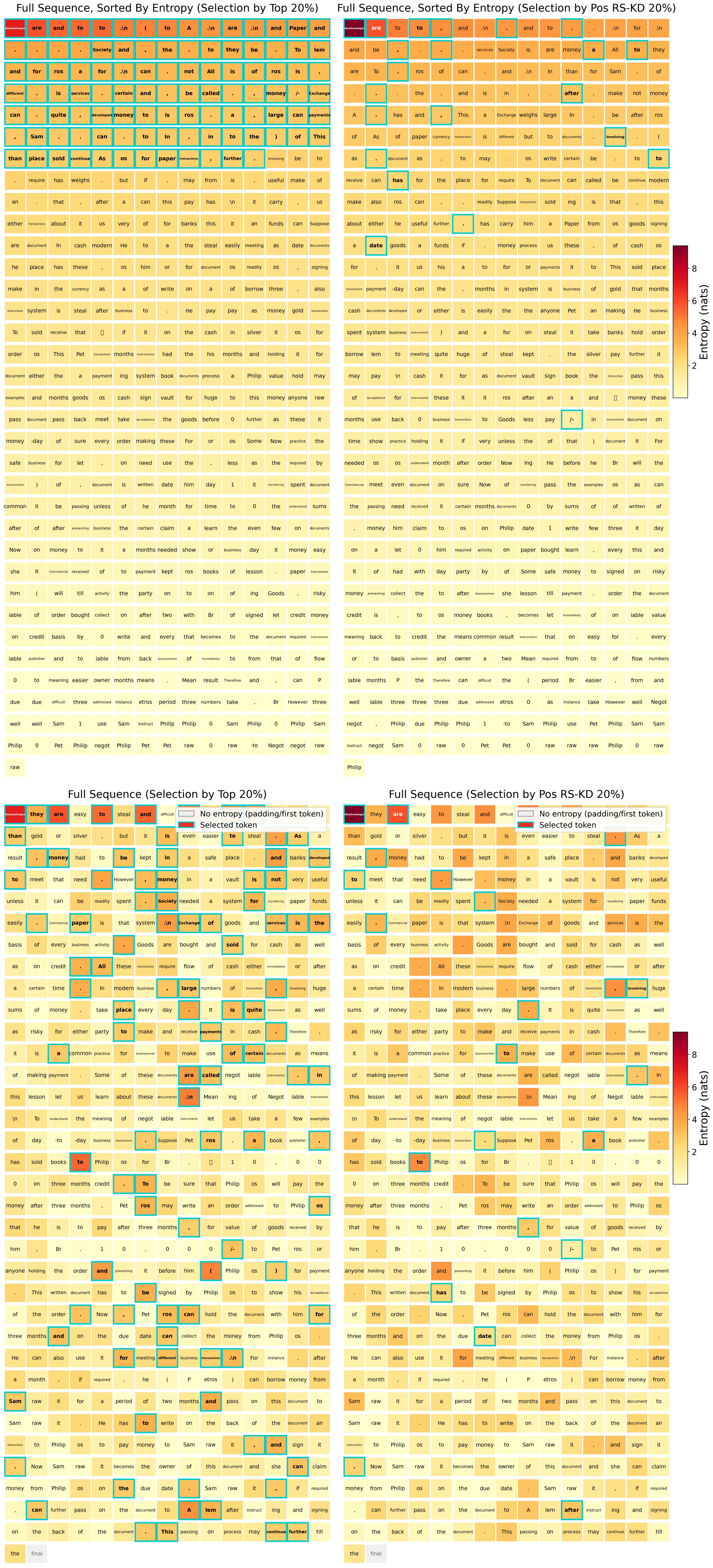}
    \caption{\textbf{Top-$k$ vs.\ positional random sampling at a fixed budget ($k{=}20\%$).}
    Tokens are colored by student entropy; teal outlines mark selected positions.
    \textbf{Top row:} the same sequence sorted by entropy, highlighting how each policy allocates its budget across the entropy distribution.
    \textbf{Bottom row:} the original token order (with padding shown as ``no entropy''), showing how selections are distributed along the sequence.
    \textbf{Left:} deterministic Top-$k$; \textbf{Right:} entropy-proportional positional RS-KD (Pos RS-KD / Pos RS-KD$^{*}$).}
    \label{fig:pos-rs-full-sorted}
\end{figure*}

Fig.~\ref{fig:pos-rs-full-sorted} visualizes the difference between deterministic Top-$k\%$ position selection and positional random sampling (Pos RS-KD) at the same budget.
While Pos RS-KD is attractive because it introduces stochasticity according to an uncertainty-derived weight, it underperformed Top-$k$ in both accuracy and calibration in our general-purpose setting (Table~\ref{tab:student-entropy-variants}).

A possible explanation is entropy-mass concentration within a sequence: if the per-sequence entropy distribution is highly peaked, then sampling proportionally to entropy can allocate a large fraction of the budget to a small set of extreme-entropy positions (often near the beginning of the sequence).This can reduce coverage of other informative positions that Top-$k$ would deterministically include, and may increase variance across updates.

There are several simple mitigations that may improve Pos RS-KD in future work:
(i) \emph{temperature smoothing} of the sampling distribution (e.g., sampling from $\propto H(q_t)^{1/T}$ with $T>1$) to flatten overly-peaked sequences;
(ii) lightweight \emph{heuristics} such as excluding the first few eligible positions or clipping extreme entropies; and
(iii) combining entropy-proportional sampling with a small deterministic ``coverage'' component (e.g., reserving part of the budget for Top-$k\%$ and sampling the remainder).
We leave a systematic study of these variants to future work.

\clearpage

\section{The Offline Cache Tradeoff}
\label{sec:offline_cache}

The cache footprint of \triaxselector{} could be further reduced by storing teacher logits only for a fixed subset of selected positions, scaling storage with the position budget \(k\%\).
However, this introduces a fundamental tradeoff: position-level caching maximizes storage savings but breaks adaptivity, whereas sample-level caching preserves curriculum effects at the cost of a larger cache.
We therefore avoid position-level caching, as our default student-entropy selector induces an implicit curriculum-high-entropy positions evolve during training, and freezing a precomputed mask would remove this adaptivity and may degrade distillation quality.

\begin{figure*}[t]
    \centering
    \includegraphics[width=0.5\linewidth]{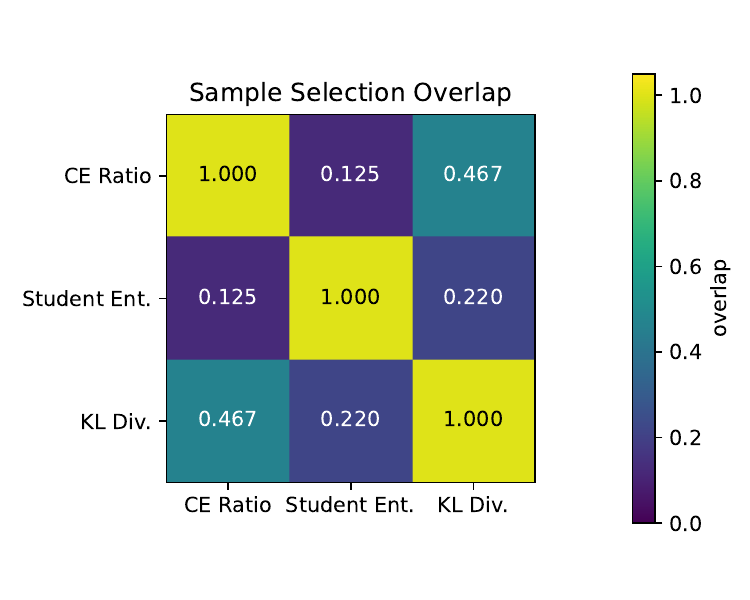}
    \caption{\textbf{Sample-selection overlap across metrics.}
Pairwise overlap between samples selected by student entropy, CE ratio, and KL divergence.
Teacher-based metrics show higher mutual overlap than with student entropy, indicating differing selection stability.}
    \label{fig:samples-skipping-overlap}
\end{figure*}

In contrast, we hypothesize that \emph{sample-level} selection is more stable under student learning dynamics, making it suitable for a one-shot prefiltering pass that reduces teacher queries and cache size.
This hypothesis is consistent with \citet{xu2020unix}, who show that samples uncertain for the student are also hard for the teacher, suggesting that sample difficulty is largely data-inherent.
However, we do not claim to establish this conclusively.
As shown in Fig.~\ref{fig:samples-skipping-overlap}, an exploratory overlap analysis reveals substantial agreement between teacher-based metrics (KL and CE ratio; 46.7\%), but much lower overlap between student entropy and either metric (22.0\% and 12.5\%), highlighting the need for a more systematic study of sample-selection stability.

An alternative is to construct the cache online during distillation, recording positions or samples selected by the evolving student.
While this preserves curriculum effects and may transfer across students, it sacrifices a key benefit of offline caching-the ability to distill while holding only one model in memory-and is less suitable for multi-epoch training.
Future work could compare samples selected under online curricula (e.g., GLS) to those from a pre-distillation pass to better characterize selection stability.

\section{Validation Split Tables}
\label{sec:appendix-val}
\FloatBarrier

This appendix reports validation-split results used for model/metric selection and ablations during development.
All comparisons in the main paper are based on the held-out test split; the validation split is not used for final evaluation.
We evaluated each validation experiment using three seeds (1337, 1338, 1339) and report the average results.
The validation benchmark suite is constructed from the average accuracy across the validation splits of ArcEasy, GSM8K, HellaSwag, and PIQA.

\begin{table}[t]
\centering
\caption{\textbf{Position-importance metrics with Top-20\% selection on the validation set}, averaged across three fixed seeds (1337, 1338, 1339), trained on 80M tokens of FineWeb-Edu. Based on these results, we selected student entropy as our position-importance metric (Table~\ref{tab:uncertainty-analysis-test}). It achieves top validation accuracy and uniquely among the top metrics, requires no teacher-side information, enabling the use of a selective LM head on the teacher that avoids logit computation at non-selected positions.}
\footnotesize
\begin{tabular}{lc}
\toprule
\textbf{Method} & \textbf{Acc. $\uparrow$} \\
\midrule
Qwen3 1.7B & 62.2 \\
Qwen3 8B & \textbf{\textit{75.1}} \\
AT-KD & 65.2 \\
Full KD & 65.6 \\
Random 20\% & 65.5 \\
\cmidrule(lr){1-2}
\multicolumn{2}{l}{\textit{Position selection policy: Top 20\%}} \vspace{3px} \\
Student entropy & \textbf{66.0} \\
Teacher entropy & 65.4 \\
Student CE & 65.8 \\
Teacher CE & 65.3 \\
KL & \textbf{66.0} \\
Reverse KL & \textbf{66.0} \\
CE ratio & 65.9 \\
CE ratio + Student Entropy & 65.8 \\
Student entropy + KL & 65.7 \\
\bottomrule
\end{tabular}
\end{table}

\begin{table}[t]
\centering
\caption{\textbf{Position-selection policies with student entropy on the validation set}, averaged across three seeds. We selected Top 20\% for our main experiments (Table~\ref{tab:student-entropy-variants}): although GLS and Curriculum achieve slightly higher validation accuracy, the differences are small (0.1--0.2 points) and Top 20\% is simpler, avoiding additional hyperparameters (queue size for GLS, schedule for Curriculum).}
\small
\begin{tabular}{lc}
\toprule
\textbf{Method} & \textbf{Acc. $\uparrow$} \\
\midrule
Qwen3 1.7B & 62.2 \\
Qwen3 8B & \textbf{\textit{75.1}} \\
AT-KD & 65.2 \\
Full KD & 65.6 \\
Random 20\% & 65.5 \\
\cmidrule(lr){1-2}
\multicolumn{2}{l}{\textit{Position selection policy: Top 20\%}} \vspace{3px} \\
Top 20\% & 66.0 \\
Curriculum Learning 20\% & 66.1 \\
GLS 30K 20\% & \textbf{66.2} \\
Pos RS-KD 20\% & 64.9 \\
Pos RS-KD$^*$ 20\% & 65.5 \\
\bottomrule
\end{tabular}
\label{tab:student-entropy-variants-validation-mean}
\end{table}

\FloatBarrier

\section{Standard Deviations}
\label{sec:std}
We report standard deviations over three fixed random seeds to quantify run-to-run variance under an otherwise identical training setup.
\begin{table}[t]
\centering
\caption{
Standard deviations for Table~\ref{tab:uncertainty-analysis-test} (position-importance metrics with Top-20\% selection), computed over three fixed seeds (1337, 1338, 1339).}
\small
\begin{tabular}{lcccc}
\toprule
\textbf{Method} & \textbf{Accuracy $\uparrow$} & \textbf{IFEval $\uparrow$} & \textbf{PPL $\downarrow$} & \textbf{ECE $\downarrow$} \\
\midrule
Full KD & 0.20 & 0.56 & 0.18 & 0.07 \\
RandomPos 20\% & 0.15 & 1.25 & 0.21 & 0.59 \\
AT-KD & 0.04 & 0.78 & 0.06 & 0.04 \\
\cmidrule(lr){1-5}
\multicolumn{5}{l}{\textit{Position selection policy: Top 20\%}} \vspace{3px} \\
Student Entropy (\selector{}) & 0.14 & 0.81 & 0.26 & 0.12 \\
Teacher Entropy & 0.68 & 0.24 & 1.28 & 0.52 \\
Teacher CE & 0.30 & 0.67 & 0.38 & 0.79 \\
Student CE & 0.16 & 0.36 & 0.35 & 0.20 \\
KL & 0.16 & 1.19 & 0.09 & 0.11 \\
Reverse KL & 0.10 & 0.45 & 0.09 & 0.04 \\
CE ratio & 0.02 & 0.12 & 0.34 & 0.06 \\
CE ratio + Student Entropy & 0.06 & 0.44 & 0.04 & 0.07 \\
Student Entropy + KL & 0.13 & 0.13 & 0.48 & 0.49 \\
\bottomrule
\end{tabular}
\label{tab:uncertainty-analysis-test-std}
\end{table}

\begin{table}[t]
\centering
\caption{Standard deviations for Table~\ref{tab:student-entropy-variants} (position-selection policies with student entropy), computed over three fixed seeds (1337,1338,1339).}
\small
\begin{tabular}{lcccc}
\toprule
\textbf{Method} & \textbf{Accuracy $\uparrow$} & \textbf{IFEval $\uparrow$} & \textbf{PPL $\downarrow$} & \textbf{ECE $\downarrow$} \\
\midrule
Full KD & 0.20 & 0.56 & 0.18 & 0.07 \\
RandomPos 20\% & 0.15 & 1.25 & 0.21 & 0.59 \\
AT-KD & 0.04 & 0.78 & 0.06 & 0.04 \\
\cmidrule(lr){1-5}
\multicolumn{5}{l}{\textit{Position importance metric: Student entropy, \(k=20\%\)}\vspace{3px}} \\
Top 20\% (\selector{}) & 0.14 & 0.81 & 0.26 & 0.12 \\
Top 20\% GLS 30K & 0.23 & 0.33 & 0.55 & 0.14 \\
Curriculum 20\% & 0.09 & 0.41 & 0.21 & 0.08 \\
Pos RS-KD$^*$ 20\% & 0.39 & 0.36 & 0.44 & 0.10 \\
Pos RS-KD 20\% & 0.08 & 0.99 & 0.26 & 0.15 \\
\bottomrule
\end{tabular}
\label{tab:student-entropy-variants-std}
\end{table}

\begin{table}[t]
\centering
\caption{Standard deviations for Table~\ref{tab:rskd-samples-comparison} (general-purpose distillation; test split, 80M tokens), computed over three fixed seeds.}
\small
\begin{tabular}{lcccc}
\toprule
\textbf{Method} & \textbf{Accuracy (\%) $\uparrow$} & \textbf{IFEval (\%) $\uparrow$} & \textbf{PPL $\downarrow$} & \textbf{ECE (\%) $\downarrow$} \\
\midrule
Full KD & 0.20 & 0.56 & 0.18 & 0.07 \\
RandomPos 20\% & 0.15 & 1.25 & 0.21 & 0.59 \\
RandomSmp 20\% & 0.27 & 0.06 & 0.71 & 0.03 \\
\selector{} & 0.13 & 0.48 & 0.26 & 0.08 \\
RS-KD & 0.06 & 5.33 & 0.06 & 0.01 \\
TopSmp 20\% & 0.58 & 0.48 & 0.64 & 0.05 \\
RS-KD + TopSmp 20\% & 0.11 & 0.21 & 0.00 & 0.04 \\
\selector{} + TopSmp 20\% & 0.07 & 0.77 & 0.27 & 0.11 \\
\triaxselector{} & 0.15 & 1.61 & 0.12 & 0.16 \\
\bottomrule
\end{tabular}
\label{tab:rskd-samples-comparison-test-std}
\end{table}

\begin{table}[t]
\centering
\caption{Standard deviations for Table~\ref{tab:gsm8k-finetune-comparison} (task-specific GSM8K distillation; test split), computed over three fixed seeds (1340, 1341, 1342).}
\small
\begin{tabular}{clcccc}
\toprule
& \textbf{Method} & \textbf{GSM8K Acc.} & \textbf{Acc.} & \textbf{PPL} \\
\midrule
\multirow{7}{*}{\rotatebox{90}{\textit{Off Policy Distill.}}}
& Full KD & 0.90 & 0.10 & 0.06 \\
& Random 20\% & 0.10 & 0.12 & 0.15 \\
\cmidrule(lr){2-5}
& \selector{} & 0.12 & 0.06 & 0.10 \\
& Pos-RS-KD 20\% & 0.80 & 0.17 & 0.23 \\
& Pos RS-KD$^*$ 20\% & 1.22 & 0.36 & 0.31 \\
& TopSmp 20\% & 0.12 & 0.00 & 0.06 \\
& \selector{} + TopSmp 20\% & 0.06 & 0.06 & 0.06 \\
& \triaxselector{} & 0.31 & 0.15 & 0.00 \\
\midrule
\multirow{8}{*}{\rotatebox{90}{\textit{On Policy Distill.}}}
& Full KD & 0.21 & 0.00 & 0.06 \\
& Random 20\% & 0.72 & 0.06 & 0.12 \\
\cmidrule(lr){2-5}
& \selector{} & 0.15 & 0.00 & 0.00 \\
& Pos-RS-KD 20\% & 2.25 & 0.00 & 0.58 \\
& Pos-RS-KD$^*$ 20\% & 0.58 & 0.06 & 0.21 \\
& TopSmp 20\% & 0.49 & 0.06 & 0.12 \\
& \selector{} + TopSmp 20\% & 1.00 & 0.31 & 0.66 \\
\bottomrule
\end{tabular}
\label{tab:gsm8k-finetune-comparison-test-std}
\end{table}
\clearpage

\section{Memory Efficiency of Selective LM Head and Chunked Streaming Entropy Computation}
\label{sec:memory_utilization}

\begin{table*}[t]
    \centering
\caption{\textbf{Memory and speed comparison of selective LM head configurations.}
Experiments use Qwen3-8B (teacher) $\rightarrow$ Qwen3-1.7B (student) on NVIDIA GeForce RTX 3090 GPUs with batch size $B{=}2$, sequence length $T{=}512$, and $\lambda{=}1$.
\emph{Default flow} corresponds to the standard KD implementation.
\emph{Chunked-streaming flow} restructures the computation to match the selective implementation (e.g., streaming entropy computation and position indexing) while selecting all positions ($k{=}100\%$), isolating the overhead of code reorganization.
At $k{=}20\%$, only the top 20\% of positions (by student entropy) participate in the KD loss.
Speedup is reported relative to the default flow baseline.}
    \label{tab:memory-comparison}
\small
\resizebox{\linewidth}{!}{%
\begin{tabular}{cllccccc}
\toprule
& \textbf{Configuration} & \textbf{$k$} & \textbf{Student Peak (GB)} & \textbf{Teacher Peak (GB)} & \textbf{Wall Time} & \textbf{Speedup} \\
\midrule

\multirow{8}{*}{\rotatebox{90}{\textit{1M tokens}}}
& Full KD \textit{(default flow)} & 100\% & 15.88 & 17.30 & 16.3 min & 1.00$\times$ \\
& Full KD \textit{(chunked flow)} & 100\% & 14.15 & 17.58 & 14.9 min & 1.09$\times$ \\
& Teacher selective LM head \textit{(chunked flow)} & 100\% & 14.15 & 17.29 & 14.9 min & 1.09$\times$ \\
\cmidrule(lr){2-7}
& No selective LM head \textit{(default flow)} & 20\% & 13.59 & 17.30 & 13.6 min & 1.20$\times$ \\
& No selective LM head \textit{(chunked flow)} & 20\% & \textbf{11.42} & 15.97 & 12.6 min & 1.29$\times$ \\
\cmidrule(lr){2-7}
& Teacher selective LM head \textit{(chunked flow)} & 20\% & \textbf{11.42} & \textbf{15.68} & 12.6 min & 1.29$\times$ \\
& Student selective LM head \textit{(chunked flow)} & 20\% & \textbf{11.42} & 15.97 & 12.2 min & 1.34$\times$ \\
& Teacher + Student selective LM head \textit{(chunked flow)} & 20\% & \textbf{11.42} & \textbf{15.68} & \textbf{12.0 min} & 1.36$\times$ \\

\midrule

\multirow{7}{*}{\rotatebox{90}{\textit{5M tokens}}}
& Full KD \textit{(default flow)} & 100\% & 15.88 & 17.30 & 79.2 min & 1.00$\times$ \\
& Full KD \textit{(chunked flow)} & 100\% & 14.15 & 17.58 & 72.0 min & 1.10$\times$ \\
\cmidrule(lr){2-7}
& No selective LM head \textit{(default flow)} & 20\% & 13.59 & 17.30 & 69.7 min & 1.14$\times$ \\
& No selective LM head \textit{(chunked flow)} & 20\% & \textbf{11.42} & 15.97 & 65.5 min & 1.21$\times$ \\
\cmidrule(lr){2-7}
& Teacher selective LM head \textit{(chunked flow)} & 20\% & \textbf{11.42} & \textbf{15.68} & 62.5 min & 1.27$\times$ \\
& Student selective LM head \textit{(chunked flow)} & 20\% & \textbf{11.42} & 15.97 & 61.8 min & 1.28$\times$ \\
& Teacher + Student selective LM head \textit{(chunked flow)} & 20\% & \textbf{11.42} & \textbf{15.68} & \textbf{60.1 min} & 1.32$\times$ \\

\midrule

\multirow{5}{*}{\rotatebox{90}{\textit{80M tokens}}}
& Full KD \textit{(default flow)} & 100\% & 15.88 & 17.30 & 21h10m & 1.00$\times$ \\
& Full KD \textit{(chunked flow)} & 100\% & 14.15 & 17.58 & 19h37m & 1.08$\times$ \\
\cmidrule(lr){2-7}
& Teacher selective LM head \textit{(chunked flow)} & 20\% & \textbf{11.42} & \textbf{15.68} & 17h04m & 1.24$\times$ \\
& Student selective LM head \textit{(chunked flow)} & 20\% & \textbf{11.42} & 15.97 & 16h57m & 1.25$\times$ \\
& Teacher + Student selective LM head \textit{(chunked flow)} & 20\% & \textbf{11.42} & \textbf{15.68} & \textbf{15h47m} & 1.34$\times$ \\

\bottomrule
\end{tabular}
}
\end{table*}

\begin{figure*}[t]
    \centering
    \includegraphics[width=\linewidth]{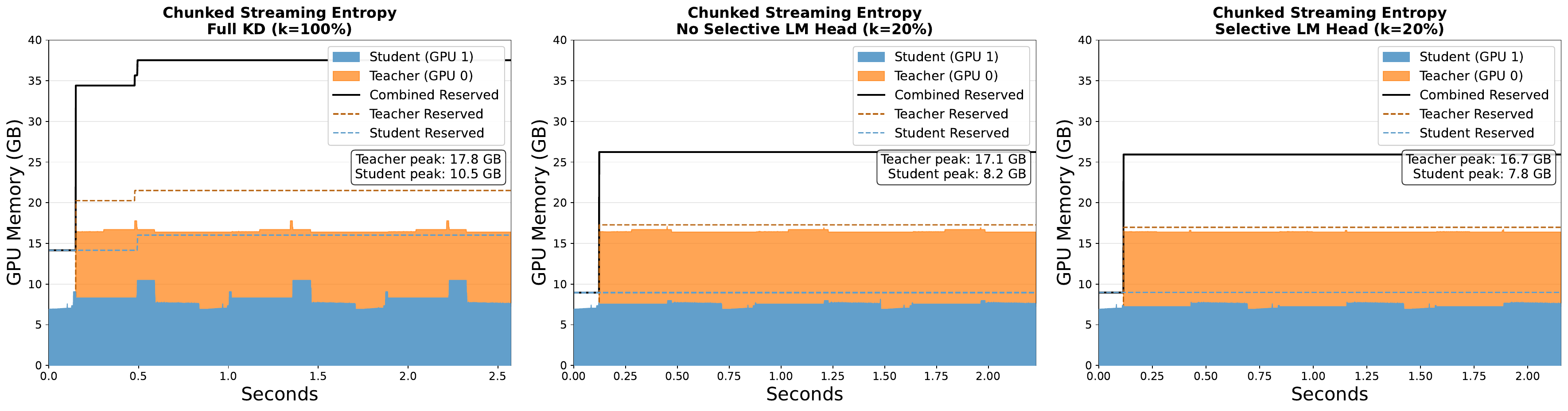}
    \caption{\textbf{GPU memory profiles under different selective LM head configurations.}
    Memory traces from PyTorch profiler over several training steps using chunked-streaming entropy computation.
    \textbf{Left:} Full KD with $k{=}100\%$, where allocating full $[B,L,V]$ logit tensors induces periodic memory spikes.
    \textbf{Middle:} $k{=}20\%$ without selective LM head, where fewer positions participate in KD but full logits are still materialized.
    \textbf{Right:} $k{=}20\%$ with selective LM head, where logits are computed only at selected positions, eliminating transient spikes and further reducing peak memory.
    Teacher and student run on separate GPUs; stacked areas show allocated memory and dashed lines indicate reserved memory.}
    \label{fig:memory-comparison}
\end{figure*}

Table~\ref{tab:memory-comparison} presents a detailed ablation of memory usage and training speed across different KD implementations.
Specifically, we compare:
\begin{enumerate}[leftmargin=*,nosep]
    \item \textbf{Default KD implementation}: Standard KD that computes full $[B,L,V]$ logits for both teacher and student.
    \item \textbf{Chunked-streaming implementation}: Incorporates chunked-streaming entropy computation and the selective code path, while still computing logits at all positions. This isolates the effect of chunked streaming independent of position sparsification.
    \item \textbf{Selective LM head variants}: Compute KD loss on a subset of positions selected by student entropy, with teacher- and/or student-side selective LM heads restricting logit computation and gradient propagation to selected positions.
\end{enumerate}

Several observations emerge from Table~\ref{tab:memory-comparison}.
First, even at $k{=}100\%$, the chunked-streaming flow reduces student peak memory from 15.88\,GB to 14.15\,GB (11\%) and yields a 9\% speedup by avoiding materialization of full student logit tensors during backpropagation.
Second, reducing $k$ from 100\% to 20\% provides substantial additional savings: even without a selective LM head, student peak memory drops to 13.59\,GB (default flow) or 11.42\,GB (chunked-streaming flow), with speedups of 1.20$\times$ and 1.29$\times$, respectively.
Third, adding a selective LM head at $k{=}20\%$ further reduces teacher peak memory from 15.97\,GB to 15.68\,GB while maintaining the same student memory footprint; the combined teacher + student selective configuration achieves the best wall time (12.0\,min, 1.36$\times$ speedup).

Fig.~\ref{fig:memory-comparison} visualizes these effects over time.
At $k{=}100\%$ (left panel), memory spikes arise from transient allocation of full $[B,L,V]$ logit tensors during each training step.
Reducing to $k{=}20\%$ without a selective LM head (middle panel) already lowers peak memory, as fewer positions participate in the KD loss, though full logits are still materialized.
With a selective LM head at $k{=}20\%$ (right panel), the spikes are eliminated entirely, as logits are computed only at the selected $\sim$20\% of positions.

\end{document}